\documentclass[final]{cvpr}

\usepackage{times}
\usepackage{epsfig}
\usepackage{graphicx}
\usepackage{amsmath}
\usepackage{amssymb}
\usepackage{caption}
\usepackage[ruled,vlined,linesnumbered]{algorithm2e}
\usepackage{enumitem}

\usepackage{makecell}
\usepackage{cases}
\usepackage{algorithmic}

\usepackage[pagebackref=true,breaklinks=true,colorlinks,bookmarks=false]{hyperref}

\def\cvprPaperID{3093} 
\def\confYear{CVPR 2021}

\begin{document}

\title{LiDAR-based Panoptic Segmentation via Dynamic Shifting Network}


\author{Fangzhou Hong$^{\dag}$~~~ Hui Zhou$^{\ddagger}$~~~ Xinge Zhu$^{\ddagger}$~~~ Hongsheng Li$^{\ddagger}$~~~ Ziwei Liu$^{\dag}$ \and {\small $^{\dag}$Nanyang Technological University~~~ $^{\ddagger}$Chinese University of Hong Kong}}


\newcommand{\nickname}{DS-Net}
\newcommand{\fullname}{Dynamic Shifting Network}
\newcommand{\things}{\textit{things}}
\newcommand{\stuff}{\textit{stuff}}
\newcommand{\PQ}{PQ}
\newcommand{\PQda}{PQ\textsuperscript{$\dagger$}}
\newcommand{\RQ}{RQ}
\newcommand{\SQ}{SQ}
\newcommand{\PQth}{PQ\textsuperscript{Th}}
\newcommand{\RQth}{RQ\textsuperscript{Th}}
\newcommand{\SQth}{SQ\textsuperscript{Th}}
\newcommand{\PQst}{PQ\textsuperscript{St}}
\newcommand{\RQst}{RQ\textsuperscript{St}}
\newcommand{\SQst}{SQ\textsuperscript{St}}
\newcommand{\miou}{mIoU}

\newcommand{\road}{\rotatebox[origin=c]{90}{road}}
\newcommand{\side}{\rotatebox[origin=c]{90}{sidewalk}}
\newcommand{\park}{\rotatebox[origin=c]{90}{parking}}
\newcommand{\ogro}{\rotatebox[origin=c]{90}{other ground}}
\newcommand{\buil}{\rotatebox[origin=c]{90}{building}}
\newcommand{\car}{\rotatebox[origin=c]{90}{car}}
\newcommand{\truc}{\rotatebox[origin=c]{90}{truck}}
\newcommand{\bcle}{\rotatebox[origin=c]{90}{bicycle}}
\newcommand{\mcle}{\rotatebox[origin=c]{90}{motorcycle}}
\newcommand{\oveh}{\rotatebox[origin=c]{90}{other-vehicle}}
\newcommand{\vege}{\rotatebox[origin=c]{90}{vegetation}}
\newcommand{\trun}{\rotatebox[origin=c]{90}{trunk}}
\newcommand{\terr}{\rotatebox[origin=c]{90}{terrain}}
\newcommand{\pers}{\rotatebox[origin=c]{90}{person}}
\newcommand{\bcli}{\rotatebox[origin=c]{90}{bicyclist}}
\newcommand{\mcli}{\rotatebox[origin=c]{90}{motorcyclist}}
\newcommand{\fenc}{\rotatebox[origin=c]{90}{fence}}
\newcommand{\pole}{\rotatebox[origin=c]{90}{pole}}
\newcommand{\traf}{\rotatebox[origin=c]{90}{traffic sign}}

\twocolumn[{
    \renewcommand\twocolumn[1][]{#1}%
    \maketitle
    \vspace{-30pt}
    \begin{center}
        \centering
        \includegraphics[width=1.0\textwidth]{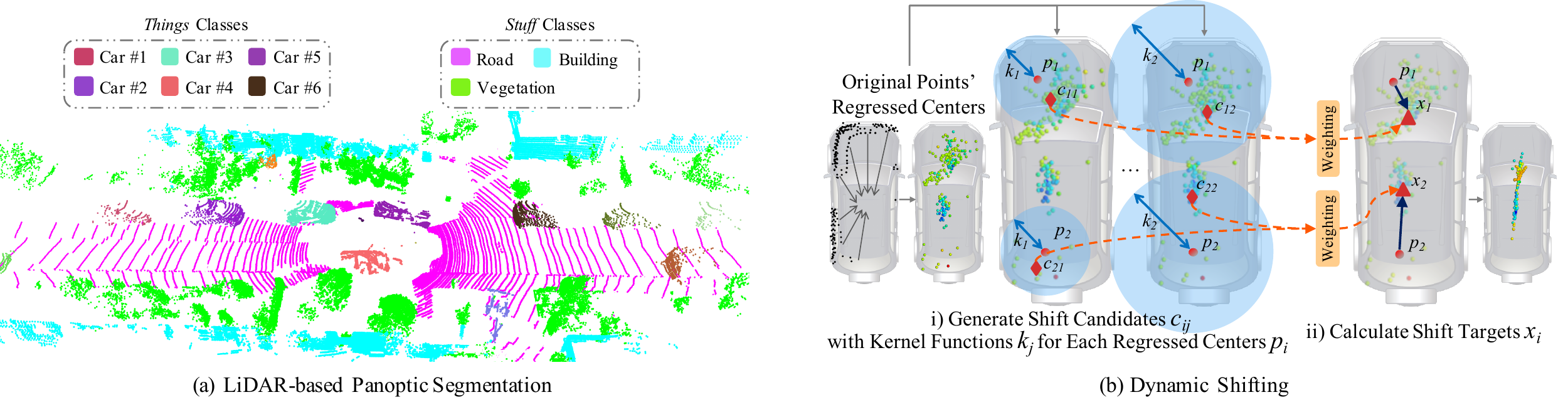}
        \vspace{-20pt}
        \captionof{figure}{As shown in (a), LiDAR-based panoptic segmentation requires instance-level segmentation for \textit{things} classes and semantic-level segmentation for \textit{stuff} classes. (b) shows the core operation of the proposed dynamic shifting where several shift candidates are weighted to obtain the optimal shift target for each regressed center.}
        \label{fig:01_teaser}
    \end{center}
}]

\begin{abstract}
With the rapid advances of autonomous driving, it becomes critical to equip its sensing system with more holistic 3D perception.
However, existing works focus on parsing either the objects (\eg cars and pedestrians) or scenes (\eg trees and buildings) from the LiDAR sensor.
In this work, we address \textbf{the task of LiDAR-based panoptic segmentation}, which aims to parse both objects and scenes in a unified manner.
As one of the first endeavors towards this new challenging task, we propose the \fullname{} (\nickname{}), which serves as an effective panoptic segmentation framework in the point cloud realm.
In particular, \nickname{} has three appealing properties:
\textbf{1) strong backbone design.} \nickname{} adopts the cylinder convolution that is specifically designed for LiDAR point clouds. 
The extracted features are shared by the semantic branch and the instance branch which operates in a bottom-up clustering style.
\textbf{2) Dynamic Shifting for complex point distributions.} We observe that commonly-used clustering algorithms like BFS or DBSCAN are incapable of
handling complex autonomous driving scenes with non-uniform point cloud distributions and varying
instance sizes.
Thus, we present an efficient learnable clustering module, dynamic shifting, which adapts kernel functions on-the-fly for different instances.
\textbf{3) Consensus-driven Fusion.} Finally, consensus-driven fusion is used to deal with the disagreement between semantic and instance predictions.
To comprehensively evaluate the performance of LiDAR-based panoptic segmentation, we construct and curate benchmarks from two large-scale autonomous driving LiDAR datasets, SemanticKITTI and nuScenes.
Extensive experiments demonstrate that our proposed \nickname{} achieves superior accuracies over current state-of-the-art methods.
Notably, we achieve 1st place on the public leaderboard of SemanticKITTI, outperforming 
2nd place by 2.6\% in terms of the PQ metric~\footnote{Accessed at 2020-11-16. Codes are available at \url{https://github.com/hongfz16/DS-Net}. Corresponding email: fangzhouhong820@gmail.com}.
\end{abstract}

\section{Introduction}
Autonomous driving, one of the most promising applications of computer vision, has achieved rapid progress in recent years.
Perception system, one of the most important modules in autonomous driving, has also attracted extensive studies in previous research works.
Admittedly, the classic tasks of 3D object detection \cite{lang2019pointpillars, shi2020pv, yan2018second} and semantic
segmentation \cite{milioto2019rangenet++, wu2018squeezeseg, zhang2020polarnet} have developed relatively mature solutions that support real-world autonomous driving prototypes.
However, there still exists a considerable gap between the existing works and the goal of holistic perception which is essential
for the challenging autonomous driving scenes.
In this work, we propose to close the gap by exploring the task of LiDAR-based panoptic segmentation, which requires full-spectrum point-level predictions.

Panoptic segmentation has been proposed in 2D detection \cite{kirillov2019panoptic} as a new vision task which unifies
semantic and instance segmentation.
Behley \etal \cite{behley2020benchmark} extend the task to LiDAR point clouds and propose the task of LiDAR-based panoptic segmentation.
As shown in Fig. \ref{fig:01_teaser} (a), this task requires to predict point-level semantic labels
for background (\stuff{}) classes (\eg road, building and vegetation),
while instance segmentation needs to be performed for foreground (\things{}) classes (\eg car, person and cyclist).

Nevertheless, the complex point distributions of LiDAR data make it difficult to perform reliable panoptic segmentation.
Most existing point cloud instance segmentation methods \cite{engelmann20203d, jiang2020pointgroup} are mainly designed for dense and uniform indoor point clouds.
Therefore, decent segmentation results can be achieved through the center regression and heuristic clustering algorithms.
However, due to the non-uniform density of LiDAR point clouds and varying sizes of instances,
the center regression fails to provide ideal point distributions for clustering.
The regressed centers usually form noisy strip distributions that vary in density and sizes.
As will be analyzed in Section \ref{3.2}, several heuristic clustering algorithms widely used in previous
works cannot provide satisfactory clustering results for the regressed centers of LiDAR point clouds.
To tackle the above mentioned technical challenges, we propose \fullname{} (\nickname{}) which is specifically designed for effective panoptic segmentation of LiDAR point clouds.

Firstly, we adopt a \textbf{strong backbone design} and provide a strong baseline for the new task.
Inspired by \cite{zhou2020cylinder3d}, the cylinder convolution is used
to efficiently extract grid-level features for each LiDAR frame in one pass which are further shared by the semantic and instance
branches.

Secondly, we present a novel \textbf{Dynamic Shifting Module} designed to cluster on the regressed centers with complex distributions
produced by the instance branch.
As illustrated in Fig. \ref{fig:01_teaser} (b), the proposed dynamic shifting module shifts the regressed centers to
the cluster centers.
The shift targets $x_i$ are adaptively computed by weighting across several shift candidates $c_{ij}$ which are
calculated through kernel functions $k_{j}$.
The special design of the module makes the \textit{shift} operation capable of dynamically adapting to the density or sizes of different
instances and therefore shows superior performance on LiDAR point clouds.
Further analysis also shows that the dynamic shifting module is robust and not sensitive to parameter settings.

Thirdly, the \textbf{Consensus-driven Fusion Module} is presented to unify the semantic and instance results
to obtain panoptic segmentation results.
The proposed consensus-driven fusion mainly solves the disagreement
caused by the class-agnostic style of instance segmentation.
The fusion module is highly efficient, thus brings negligible computation overhead.

Extensive experiments on SemanticKITTI demonstrate the effectiveness of our proposed \nickname{}.
To further illustrate the generalizability of \nickname{}, we customize a LiDAR-based panoptic
segmentation dataset based on nuScenes.
As one of the first works for this new task, we present several strong baseline results by combining
the state-of-the-art semantic segmentation and detection methods.
\nickname{} outperforms all the state-of-the-art methods on both benchmarks (1st place on the public leaderboard of SemanticKITTI).

The main contributions are summarized below: 
\textbf{1)} To our best knowledge, we present one of the first attempts to address the challenging task of LiDAR-based panoptic segmentation.
\textbf{2)} The proposed \nickname{} effectively handles the complex distributions of LiDAR point clouds, and achieves state-of-the-art performance on SemanticKITTI and nuScenes.
\textbf{3)} Extensive experiments are performed on large-scale datasets. We adapt existing methods to this new task for in-depth comparisons. Further statistical analyses are carried out to provide valuable observations.


\begin{figure*}[t]
    \vspace{-0.6cm}
    \begin{center}
        \includegraphics[width=1.0\linewidth]{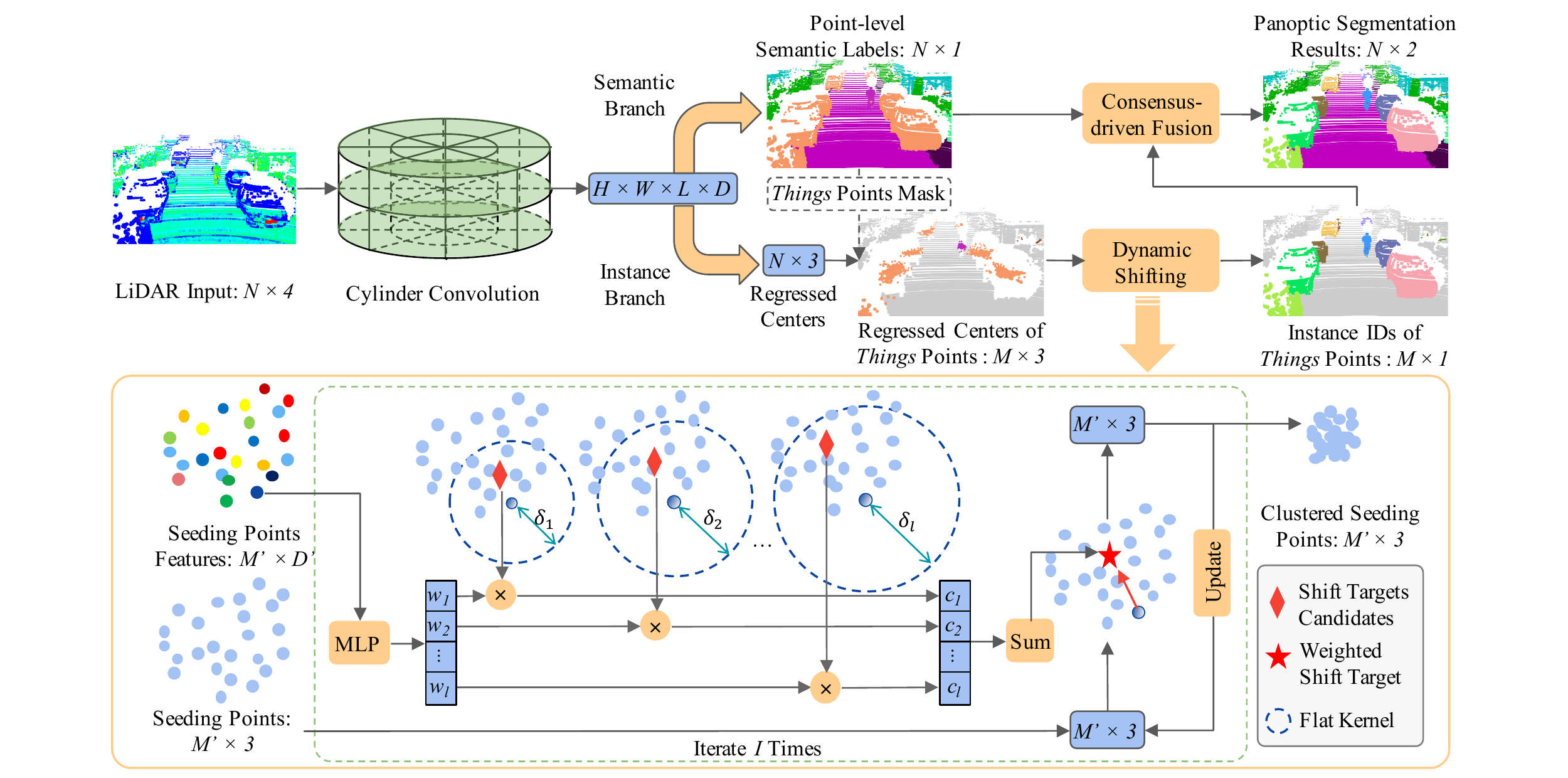}
    \end{center}
    \vspace{-0.6cm}
    \caption{\textbf{Architecture of the \nickname{}.} The \nickname{} consists of the cylinder convolution, a semantic and an instance branch as shown in the upper part of the figure. The regressed centers provided by the instance branch are clustered by the novel dynamic shifting module which is shown in the bottom half. The consensus-driven fusion module unifies the semantic and instance results into the final panoptic segmentation results.}
    \label{fig:03_dsnet_arch}
    \vspace{-0.5cm}
\end{figure*}

\section{Related Works}
\noindent\textbf{Point Cloud Semantic Segmentation.}
According to the data representations of point clouds, most point cloud semantic segmentation methods can be
categorized to point-based and voxel-based methods.
Based on PointNet \cite{qi2017pointnet} and PointNet++ \cite{qi2017pointnet++}, KPConv \cite{thomas2019kpconv},
DGCNN \cite{wang2019dynamic}, PointConv \cite{wu2019pointconv} and Randla-Net \cite{hu2020randla} can directly operate
on unordered point clouds.
However, due to space and time complexity, most point-based methods struggle on large-scale point clouds datasets
\eg ScanNet \cite{dai2017scannet}, S3DIS \cite{armeni20163d}, and SemanticKITTI \cite{behley2019semantickitti}.
MinkowskiNet \cite{choy20194d} utilizes the sparse convolutions to efficiently perform semantic segmentation on the voxelized
large-scale point clouds.
Different from indoor RGB-D reconstruct point clouds, LiDAR point clouds have non-uniform and sparse
point distributions which require special designs of the network.
SqueezeSeg \cite{wu2018squeezeseg} views LiDAR point clouds as range images while PolarNet \cite{zhang2020polarnet} and Cylinder3D \cite{zhou2020cylinder3d} divide the LiDAR point clouds under the polar and cylindrical coordinate systems.
Compared to other data representations, cylindrical division is more suitable for sparse and non-uniform LiDAR
point clouds because it
fully utilizes the localization of point clouds while remains high efficiency.

\noindent\textbf{Point Cloud Instance Segmentation.}
Previous works have shown great progress in the instance segmentation of indoor point clouds.
A large number of point-based methods (\eg SGPN \cite{wang2018sgpn}, ASIS \cite{wang2019associatively}, JSIS3D \cite{pham2019jsis3d} and JSNet \cite{zhao2020jsnet}) split the whole scene into
small blocks and learn point-wise embeddings for final clustering.
Although great efforts have been made on semantic and instance branch merging in order to boost each
other's performance, these methods are limited by the heuristic post processing steps and the lack of perception.
To avoid the problems, recent works (\eg PointGroup \cite{jiang2020pointgroup}, 3D-MPA \cite{engelmann20203d}, OccuSeg \cite{han2020occuseg}) abandon the block partition
and use sparse convolutions to extract features of the whole scene in one pass.
As for LiDAR point clouds, there are a few previous works \cite{hu2020randla, Wong2019IdentifyingUI, wu2018squeezeseg, 2020LiDARSeg} trying to tackle the problem.
Wu \etal \cite{wu2018squeezeseg} directly cluster on XYZ coordinates after semantic segmentation.
Wong \etal \cite{Wong2019IdentifyingUI} builds the network in a top-down style and incorporates metric learning.
Zhang \etal \cite{2020LiDARSeg} uses grid-level center voting to cluster points of interest in autonomous driving
scenes.
Although our work is not targeting the instance segmentation for LiDAR point clouds, the proposed \nickname{} can provide some insights into the challenging task.


\section{Our Approach}
As one of the first attempts on the task of LiDAR-based panoptic segmentation, we first introduce a strong backbone
to establish a simple baseline (Sec. \ref{3.1}), based on which two modules are further proposed.
The novel dynamic shifting module is presented to tackle the challenge of the non-uniform LiDAR point clouds distributions (Sec. \ref{3.2}).
The efficient consensus-driven fusion module combines the semantic and instance predictions and produces panoptic segmentation results (Sec. \ref{3.3}).
The whole pipeline of the \nickname{} is illustrated in Fig. \ref{fig:03_dsnet_arch}.

\subsection{Strong Backbone Design}\label{3.1}


To obtain panoptic segmentation results, it is natural to solve two sub-tasks separately, which are semantic and instance segmentation, and combine the
results.
As shown in the upper part of Fig. \ref{fig:03_dsnet_arch}, the strong backbone consists of three parts:
the cylinder convolution, a semantic branch, and an instance branch.
High quality grid-level features are extracted by the cylinder
convolution from raw LiDAR point clouds and then shared by semantic and instance branches.

\noindent\textbf{Cylinder Convolution.}
Considering the difficulty presented by the task, we find that the cylinder convolution \cite{zhou2020cylinder3d} best
meets the strict requirements of high efficiency, high performance and fully mining of 3D positional relationship.
The cylindrical voxel partition can produce more even point distribution than normal Cartesian voxel
partition and therefore leads to higher feature extraction efficiency and higher performance.
Cylindrical voxel representation combined with sparse convolutions can naturally retain and fully explore
3D positional relationship.
Thus we choose the cylinder convolution as our feature extractor.


\noindent\textbf{Semantic Branch.}
The semantic branch performs semantic segmentation by connecting MLP to the cylinder convolution to predict semantic confidences for each voxel grid.
Then the point-wise semantic labels are copied from their corresponding grids.
We use the weighted cross entropy and Lovasz Loss \cite{berman2018lovasz} as the loss function of the semantic branch.

\noindent\textbf{Instance Branch.}
The instance branch utilizes center regression to prepare the \things{} points for further clustering.
The center regression module uses MLP to adapt cylinder convolution features and make \things{}
points to regress the centers of their instances by predicting the offset vectors $O \in \mathbb{R}^{M\times 3}$
pointing from the points $P \in \mathbb{R}^{M \times 3}$ to the instance centers $C_{gt} \in \mathbb{R}^{M\times 3}$.
The loss function for instance branch can be formulated as:
\begin{equation}
    L_{ins} = \frac{1}{M}\sum_{i=0}^{M}\lVert O[i] - (C_{gt}[i] - P[i]) \rVert_1 \text{,}
\end{equation}
where $M$ is the number of \things{} points.
The regressed centers $O + P$ are further clustered to obtain the instance IDs, which can be achieved
by either heuristic clustering algorithms or the proposed dynamic shifting module which are further introduced and
analyzed in the following section.

\subsection{Dynamic Shifting}\label{3.2}

\begin{figure}[t]
    \begin{center}
        \includegraphics[width=1.0\linewidth]{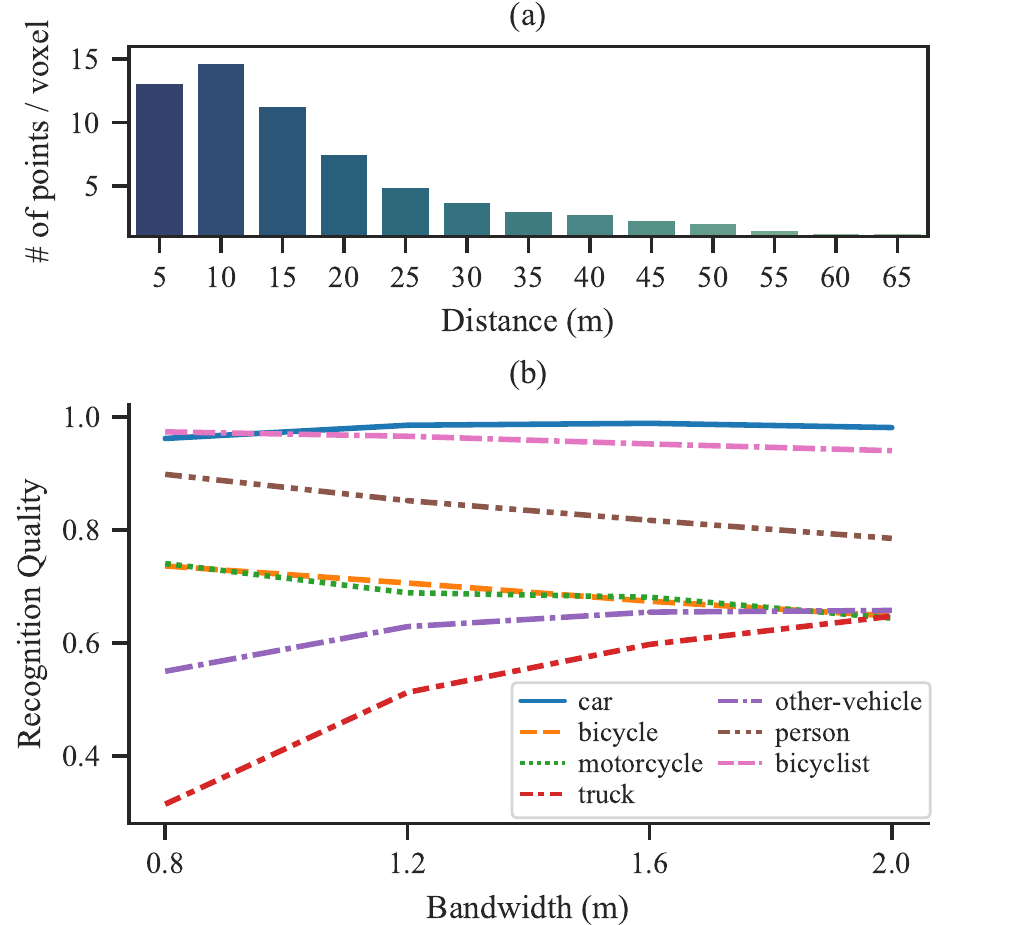}
    \end{center}
    \vspace{-0.7cm}
    \caption{(a) counts the average number of regressed centers inside each valid voxel of instances at different distances. (b) shows the effect of Different Mean Shift Bandwidth on the Recognition Quality of Different Classes.}
    \label{fig:03_02_density_and_msbandwidth}
\end{figure}

\noindent\textbf{Point Clustering Revisit.}
Unlike indoor point clouds which are carefully reconstructed using RGB-D videos, the LiDAR point
clouds have the distributions that are not suitable for normal clustering solutions used by indoor instance segmentation methods.
The varying instance sizes, the sparsity and incompleteness of LiDAR point clouds make it difficult for the center
regression module to predict the precise center location and would result in noisy long strips distribution as displayed in
Fig. \ref{fig:01_teaser} (b) instead of an ideal ball-shaped cluster around the center.
Moreover, as presented in Fig. \ref{fig:03_02_density_and_msbandwidth} (a), the clusters formed by regressed centers that are far from the LiDAR sensor
have much lower densities than those of nearby clusters due to the non-consistent sparsity of LiDAR point clouds.
Facing the non-uniform distribution of regressed centers, heuristic clustering algorithms struggle to produce satisfactory results.
Four major heuristic clustering algorithms that are used in previous bottom-up indoor point clouds instance segmentation methods
are analyzed below.
The details of the following algorithms can be found in supplementary materials.

\begin{itemize}[leftmargin=*]
    \item \textbf{Breadth First Search (BFS).}
        BFS is simple and good enough for indoor point clouds as proved in \cite{jiang2020pointgroup}, but not suitable
        for LiDAR point clouds.
        As discussed above, large density difference between clusters means that the \emph{fixed radius}
        cannot properly adapt to different clusters.
        Small radius will over-segment distant instances while large radius will under-segment near instances.
    \item \textbf{DBSCAN \cite{ester1996density} and HDBSCAN \cite{campello2013density}.}
        As density-based clustering algorithms, there is no surprise that these two algorithms also perform badly on the
        LiDAR point clouds, even though they are proved to be effective for clustering indoor point clouds
        \cite{engelmann20203d, zhang2020spatial}.
        The core operation of DBSCAN is the same as that of BFS.
        While HDBSCAN intuitively assumes that the points with lower density are more likely to be noise points which is
        not the case in LiDAR points.
    \item \textbf{Mean Shift \cite{comaniciu2002mean}.}
        The advantage of Mean Shift, which is used by \cite{lahoud20193d} to cluster indoor point clouds, is that the
        kernel function is not sensitive to density changes and robust to noise points which makes it more suitable than
        density-based algorithms.
        However, the \textit{bandwidth} of the kernel function has great impact on the clustering results as shown in Fig.
        \ref{fig:03_02_density_and_msbandwidth} (b).
        The fixed bandwidth cannot handle the situation of large and small instances simultaneously which makes Mean Shift
        also not the ideal choice for this task.
\end{itemize}

\begin{algorithm}[ht]
    \SetAlgoLined
    \KwIn{\textit{Things} Points $P \in \mathbb{R}^{M\times 3}$, \textit{Things} Features $F\in\mathbb{R}^{M\times D'}$, \textit{Things} Regressed Centers $C\in\mathbb{R}^{M\times 3}$, Fixed number of iteration $I\in\mathbb{N}$, Bandwidth candidates list $L\in\mathbb{R}^{l}$}
    \KwOut{Instance IDs of \textit{things} points $R\in\mathbb{R}^{M\times 1}$}
    $mask = FPS(P)$, $P' = P[mask]$ \label{algo:1} \\
    $X = C[mask]$, $F' = F[mask]$ \label{algo:2} \\
    \For{$i \gets 1\ \KwTo\ I$} { \label{algo:3}
        $W_i = Softmax(MLP(F'))$\\
        $acc = zeros\_like(X)$\\
        \For{$j \gets 1\ \KwTo\ l$} {
            $K_{ij} = (X X^T \leq L[j])$\\
            $D_{ij} = diag(K_{ij} \mathbf{1})$\\
            $acc = acc + W_i[:, j] \odot (D_{ij}^{-1}K_{ij}X)$
        }
        $X = acc$\\
    } \label{algo:12}
    $R' = cluster(X)$ \label{algo:13} \\
    $index = nearest\_neighbour(P, P')$ \label{algo:14} \\
    $R = R'[index]$ \label{algo:15} \\
    \Return $R$
    \caption{Forward Pass of the Dynamic Shifting Module}
    \label{algo:forward_pass}
\end{algorithm}
\setlength{\textfloatsep}{6pt}

\noindent\textbf{Dynamic Shifting.}
%
As discussed above, it is a robust way of estimating cluster centers of regressed centers by iteratively
applying kernel functions as in Mean Shift.
However, the fixed bandwidth of kernel functions fails to adapt to varying instance sizes.
Therefore, we propose the dynamic shifting module which can automatically adapt the kernel function for
each LiDAR point in the complex autonomous driving scene so that the regressed centers can be dynamically,
efficiently and precisely shifted to the correct cluster centers.

In order to make the kernel function learnable, we first consider how to mathematically define a differentiable
\textit{shift} opration.
Inspired by \cite{kong2018recurrent}, the shift operation on the seeding points (\ie points to be clustered)
can be expressed as matrix operations if the number of iterations is fixed.
Specifically, one iteration of shift operation can be formulated as follows.
Denoting $X\in \mathbb{R}^{M\times 3}$ as the $M$ seeding points, $X$ will be updated once
by the shift vector $S \in \mathbb{R}^{M\times 3}$ which is formulated as
\begin{equation}
    X \leftarrow X + \eta S \text{,}
\end{equation}
where $\eta$ is a scaling factor which is set to $1$ in our experiments.

The calculation of the shift vector $S$ is by applying kernel function $f$ on $X$, and formally defined as
$S = f(X) - X$.

Among various kinds of kernel functions, the flat kernel is simple but effective for generating shift target
estimations for LiDAR points, which is introduced as follows.
The process of applying flat kernel can be thought of as placing a query ball of certain radius (\ie bandwidth) centered at each seeding
point and the result of the flat kernel is the mass of the points inside the query ball.
Mathematically, the flat kernel $f(X)=D^{-1} K X$ is defined by the
kernel matrix $K=(X X^T \leq \delta)$, which masks out the points within a certain bandwidth $\delta$ for each seeding point,
and the diagonal matrix $D=diag(K \mathbf{1})$ that represents the number of points within the seeding point's bandwidth.


With a differentiable version of the shift operation defined, we proceed to our goal
of dynamic shifting by adapting the kernel function for each point.
In order to make the kernel function adaptable for instances with different sizes, the optimal bandwidth
for each seeding point has to be inferenced dynamically.
A natural solution is to directly regress bandwidth for each seeding point, which however is not differentiable
if used with the flat kernel.
Even though Gaussian kernel can make direct bandwidth regression trainable, it is still not the best solution
as analyzed in section \ref{ablation}.
Therefore, we apply the design of weighting across several bandwidth candidates to dynamically adapt to the optimal one.

One iteration of dynamic shifting is formally defined as follows.
As shown in the bottom half of Fig. \ref{fig:03_dsnet_arch}, $l$ bandwidth candidates
$L=\{\delta_1, \delta_2, ...,\delta_l\}$ are set.
For each seeding point, $l$ shift target candidates are calculated by $l$ flat kernels with corresponding
bandwidth candidates.
Seeding points then dynamically decide the final shift targets, which are ideally the closest to the cluster centers,
by learning the weights $W \in \mathbb{R}^{M\times l}$ to weight on $l$ candidate targets.
The weights $W$ are learned by applying MLP and Softmax on the backbone features so that
$\sum_{j=1}^{l}W[:, j] = \textbf{1}$.
The above procedure and the new learnable kernel function $\hat{f}$ can be formulated as
\begin{equation}
    \hat{f}(X) = \sum_{j=1}^{l} W[:, j] \odot (D_j^{-1} K_j X) \text{,}
\end{equation}
where $K_j = (X X^T \leq \delta_j)$ and $D_j = diag(K_j \mathbf{1})$.

With the one iteration of dynamic shifting stated clearly, the full pipeline of the dynamic shifting module, which is
formally defined in algorithm \ref{algo:forward_pass}, can be illustrated as follows.
Firstly, to maintain the efficiency of the algorithm, farthest point sampling (FPS) is performed on $M$ \things{}
points to provide $M'$ seeding points for the dynamic shifting iterations (Lines \ref{algo:1}--\ref{algo:2}).
After a fixed number $I$ of dynamic shifting iterations (Lines \ref{algo:3}--\ref{algo:12}), all seeding points have converged to the cluster centers.
A simple heuristic clustering algorithm is performed to cluster the converged seeding points to obtain
instance IDs for each seeding point (Line \ref{algo:13}).
Finally, all other \things{} points find the nearest seeding points and the corresponding instance IDs are
assigned to them (Lines \ref{algo:14}--\ref{algo:15}).

The optimization of dynamic shifting module is not intuitive since it is impractical to obtain the ground
truth bandwidth for each seeding point.
The loss function has to encourage seeding points shifting towards their cluster
centers which have no ground truths but can be approximated by the ground truth centers of instances
$C_{gt}' \in \mathbb{R}^{M' \times 3}$.
Therefore, the loss function for the $i$th iteration of dynamic shifting is defined by the manhattan distance between
the ground truth centers $C_{gt}'$ and the $i$th dynamically calculated shift targets $X_{i}$, which can be
formulated as
\begin{equation}
    l_{i} = \frac{1}{M'} \sum_{x=1}^{M'}\lVert X_{i}[x] - C_{gt}'[x] \rVert_1\text{.}
\end{equation}

Adding up all the losses of $I$ iterations gives us the loss function $L_{ds}$ for the dynamic shifting module:
\begin{equation}
    L_{ds} = \sum_{i=1}^{I} w_{i} l_{i}\text{,}
\end{equation}
where $w_i$ are weights for losses of different iterations and are all set to $1$ in our experiments.

\begin{figure*}[ht]
    \vspace{-0.6cm}
    \begin{center}
        \includegraphics[width=1.0\linewidth]{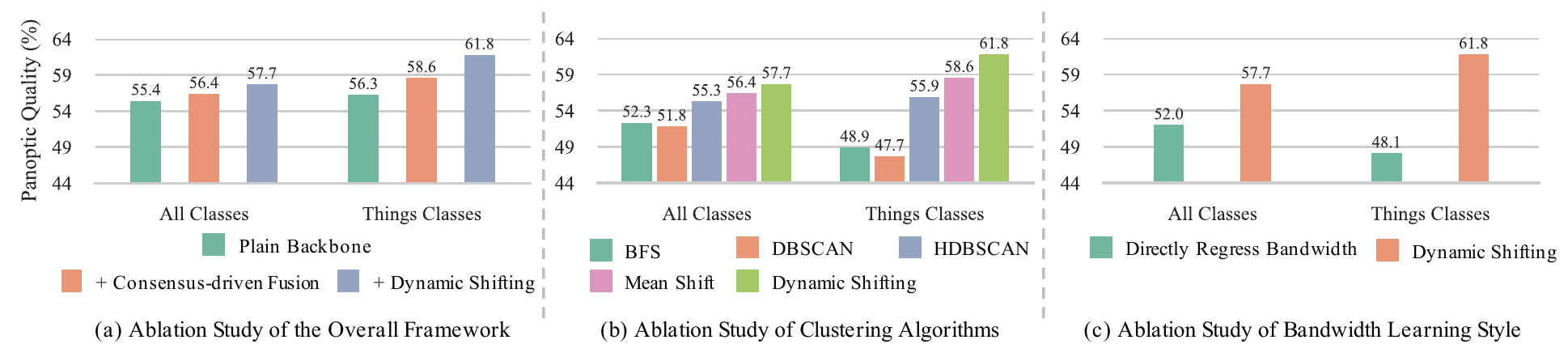}
    \end{center}
    \vspace{-0.6cm}
    \caption{\textbf{Ablation Study on the Validation Set of SemanticKITTI}. The proposed two modules both contributes to the final performance of the \nickname{}. The dynamic shifting module has advantages in clustering LiDAR point clouds. Weighting on bandwidth candidates is better than directly regressing bandwidth.}
    \label{fig:04_02_ablation}
    \vspace{-0.5cm}
\end{figure*}

\subsection{Consensus-driven Fusion} \label{3.3}
Typically, solving the conflict between semantic and instance predictions is one of the essential steps
in panoptic segmentation.
The advantages of bottom-up methods are that all points with predicted instance IDs must be in
\things{} classes and one point will not be assigned to two instances.
The only conflict needs to be solved is the disagreement of semantic predictions inside one instance,
which is brought in by the class-agnostic way of instance segmentation.
The strategy used in the proposed consensus-driven fusion is \textit{majority voting}.
For each predicted instance, the most appeared semantic label of its points determines
the semantic labels for all the points inside the instance.
This simple fusion strategy is not only efficient but could also revise and unify semantic predictions
using instance information.

\section{Experiments}
\begin{table*}[ht]
    \vspace{-0.2cm}
    \caption{LiDAR-based panoptic segmentation results on the validation set of SemanticKITTI. All results in [\%].}
    \vspace{-0.6cm}
    \begin{center}
    \small{
        \begin{tabular}{l|cccc|ccc|ccc|c}
            \Xhline{1pt}
            Method & \PQ & \PQda & \RQ & \SQ & \PQth & \RQth & \SQth & \PQst & \RQst & \SQst & \miou \\
            \hline\hline
            KPConv \cite{thomas2019kpconv} +
            PV-RCNN \cite{shi2020pv}              & 51.7         & 57.4          & 63.1          & \textbf{78.9} & 46.8          & 56.8          & \textbf{81.5} & \textbf{55.2} & \textbf{67.8} & \textbf{77.1} & 63.1          \\
            PointGroup \cite{jiang2020pointgroup} & 46.1         & 54.0          & 56.6          & 74.6          & 47.7          & 55.9          & 73.8          & 45.0          & 57.1          & 75.1                      & 55.7          \\
            LPASD \cite{milioto2020iros}             & 36.5 & 46.1 & -    & -    & -    & 28.2 & -    & -    & -    & -    & 50.7 \\
            \hline
            \nickname{}                          & \textbf{57.7} & \textbf{63.4} & \textbf{68.0} & 77.6          & \textbf{61.8} & \textbf{68.8} & 78.2          & 54.8          & 67.3          & \textbf{77.1} & \textbf{63.5} \\
            \Xhline{1pt}
        \end{tabular}
    }
    \end{center}
    \label{tab:semkitti_val}
    \vspace{-0.4cm}
\end{table*}

\begin{table*}[ht]
    \vspace{-0.2cm}
    \caption{LiDAR-based panoptic segmentation results on the test set of SemanticKITTI. All results in [\%]. ``$\ast$'' denotes the unpublished method which is in the 2nd place on the public benchmark of SemanticKITTI (accessed on 2020-11-16).}
    \vspace{-0.6cm}
    \begin{center}
    \small{
        \begin{tabular}{l|cccc|ccc|ccc|c}
            \Xhline{1pt}
            Method & \PQ & \PQda & \RQ & \SQ & \PQth & \RQth & \SQth & \PQst & \RQst & \SQst & \miou \\
            \hline\hline
            KPConv \cite{thomas2019kpconv} +
            PointPillars \cite{lang2019pointpillars} & 44.5          & 52.5          & 54.4          & 80.0          & 32.7          & 38.7          & 81.5          & 53.1          & 65.9          & 79.0                  & 58.8          \\
            RangeNet++ \cite{milioto2019rangenet++} +
            PointPillars \cite{lang2019pointpillars} & 37.1          & 45.9          & 47.0          & 75.9          & 20.2          & 25.2          & 75.2          & 49.3          & 62.8          & 76.5                  & 52.4          \\
            KPConv \cite{thomas2019kpconv} +
            PV-RCNN \cite{shi2020pv}                 & 50.2          & 57.5          & 61.4          & 80.0          & 43.2          & 51.4          & 80.2          & 55.9 & 68.7 & \textbf{79.9} & \textbf{62.8} \\
            LPASD \cite{milioto2020iros}              & 38.0          & 47.0          & 48.2          & 76.5          & 25.6          & 31.8          & 76.8          & 47.1          & 60.1          & 76.2                  & 50.9 \\
            \hline
            PolarNet\_seg$^{\ast}$ & 53.3 & 59.8 & 64.2 & 81.1 & 52.1 & 59.5 & 86.9 & 54.2 & 67.6 & 76.9 & 58.9 \\
            \hline
            \nickname{} & \textbf{55.9} & \textbf{62.5} & \textbf{66.7} & \textbf{82.3} & \textbf{55.1} & \textbf{62.8} & \textbf{87.2} & \textbf{56.5} & \textbf{69.5} & 78.7 & 61.6 \\
            \Xhline{1pt}
        \end{tabular}
    }
    \end{center}
    \label{tab:semkitti_test}
    \vspace{-0.8cm}
\end{table*}


We conduct experiments on two large-scale datasets: SemanticKITTI \cite{behley2019semantickitti} and
nuScenes \cite{nuscenes2019}.

\noindent\textbf{SemanticKITTI.}
SemanticKITTI is the first dataset that presents the challenge of LiDAR-based panoptic segmentation
and provides the benchmark\cite{behley2020benchmark}.
SemanticKITTI contains 23,201 frames for training and 20,351 frames for testing.
There are 28 annotated semantic classes which are remapped to 19 classes for the LiDAR-based panoptic segmentation task,
among which 8 classes are \things{} classes, and 11 classes are \stuff{} classes.
Each point is labeled with a semantic label and an instance id which will be set to 0 if the point belongs to
\stuff{} classes.

\noindent\textbf{nuScenes.}
In order to demonstrate the generalizability of \nickname{}, we construct another LiDAR-based panoptic
segmentation dataset from nuScenes.
With the point-level semantic labels from the newly released nuScenes \textit{lidarseg} challenge and the bounding
boxes provided by the detection task, we could generate instance labels by assigning instance IDs to
points inside bounding boxes.
The simple strategy makes good enough panoptic segmentation annotations because in the
autonomous driving scene, it is rare that something does not belong to an instance invades its bounding
box.
Following the definition of the nuScenes \textit{lidarseg} challenge, we mark 10 foreground classes as \things{} classes
and 6 background classes as \stuff{} classes out of all 16 semantic classes.
Due to the annotation of the test split of nuScenes \textit{lidarseg} challenge is held out, we could only validate
and compare the results on the validation split.
The training set of nuScenes has 28,130 frames and the validation set has 6,019 frames.

\noindent\textbf{Evaluation Metrics.}
As defined in \cite{behley2020benchmark}, the evaluation metrics of LiDAR-based panoptic segmentation are the
same as that of image panoptic segmentation defined in \cite{kirillov2019panoptic} including Panoptic Quality
(PQ), Segmentation Quality (SQ) and Recognition Quality (RQ) which are calculated across all classes.
The above three metrics are also calculated separately on \things{} and \stuff{} classes which give
\PQth{}, \SQth{}, \RQth{}, and \PQst{}, \SQst{}, \RQst{}.
\PQda{} is defined by swapping \PQ{} of each \stuff{} class to its IoU then averaging over all classes.
In addition, mean IoU (mIoU) is also used to evaluate the quality of the sub-task of
semantic segmentation.


\subsection{Ablation Study}\label{ablation}


\noindent\textbf{Ablation on Overall Framework.}
To study on the effectiveness of the proposed modules, we sequentially add consensus-driven fusion
module and dynamic shifting module to the bare backbone.
The corresponding \PQ{} and \PQth{} are reported in Fig. \ref{fig:04_02_ablation} (a) which shows
that both modules contribute to the performance of \nickname{}.
The consensus-driven fusion improves the overall performance through unifying results from two parallel branches.
The novel dynamic shifting module mainly boosts the performance of instance segmentation which are indicated by
\PQth{} where the \nickname{} outperforms our backbone (with fusion module) by 3.2\% in validation split.

\noindent\textbf{Ablation on Clustering Algorithms.}
In order to validate our previous analyses of clustering algorithms, we swap the dynamic shifting module for four
other widely-used heuristic clustering algorithms: BFS, DBSCAN, HDBSCAN, and Mean Shift.
The results are shown in Fig. \ref{fig:04_02_ablation} (b).
Consistent with our analyses in Sec. \ref{3.2}, the density-based clustering algorithms (\eg BFS, DBSCAN, HDBSCAN)
perform badly in terms of \PQ{} and \PQth{} while Mean Shift leads to the best results among the heuristic algorithms.
Moreover, our dynamic shifting module shows the superiority over all four heuristic clustering algorithms.

\noindent\textbf{Ablation on Bandwidth Learning Style.}
In the dynamic shifting module, it is natural to directly regress bandwidth for each point
instead of learning weights for several candidates, as mentioned in Sec. \ref{3.2}.
However, as shown in the Fig. \ref{fig:04_02_ablation} (c), direct regression is hard to optimize in this
case because the learning target is not straightforward.
It is difficult to determine the best bandwidth for each point, and therefore impractical to directly apply supervision
on the regressed bandwidth.
Instead, we could only determine whether the \textit{shift} takes the seeding point closer to the target
center.
Therefore, it is easier for the network to choose from and combine several bandwidth candidates.


\subsection{Evaluation Comparisons on SemanticKITTI}

\noindent\textbf{Comparison Methods.}
Since its one of the first attempts on LiDAR-based panoptic segmentation,
we provide several strong baseline results in order to validate the effectiveness of \nickname{}.
As proposed in \cite{behley2020benchmark}, one good way of constructing strong baselines is to take the results from
semantic segmentation methods and detection methods, and generate panoptic segmentation results by assigning instance
IDs to all points inside predicted bounding boxes.
\cite{behley2020benchmark} has provided the combinations of KPConv \cite{thomas2019kpconv} +
PointPillars \cite{lang2019pointpillars}, and RangeNet++ \cite{milioto2019rangenet++} +
PointPillars \cite{lang2019pointpillars}.
To make the baseline stronger, we combine KPConv \cite{thomas2019kpconv} with
PV-RCNN \cite{shi2020pv} which is the state-of-the-art 3D detection method.
In addition to the above baselines, we also adapt the state-of-the-art indoor
instance segmentation method PointGroup \cite{jiang2020pointgroup} using the official released codes to experiment on SemanticKITTI.
Moreover, LPASD \cite{milioto2020iros}, which is one of the earliest works in this area, is also included for comparison.

\noindent\textbf{Evaluation Results.}
Table \ref{tab:semkitti_val} and \ref{tab:semkitti_test} shows that the \nickname{}
outperforms all baseline methods in both validation and test splits by a large margin.
The \nickname{} surpasses the best baseline method KPConv + PV-RCNN in most metrics and especially
has the advantage of 6\% and 15\% in terms of \PQ{} and \PQth{} in validation split.
In test split, the \nickname{} outperforms KPConv + PV-RCNN by 5.7\% and 11.9\% in
\PQ{} and \PQth{}.
On the leaderboard provided by \cite{behley2020benchmark}, our \nickname{} achieves 1st place and
surpasses 2nd method ``PolarNet\_seg'' by 2.6\% and 3.0\% in \PQ{} and \PQth{} respectively.
It is worth noting that PointGroup \cite{jiang2020pointgroup} performs poorly on the LiDAR point
clouds which shows that indoor solutions are not suitable for challenging LiDAR point clouds.
Further detailed results on SemanticKITTI can be found in supplementary materials.

\begin{table*}[ht]
    \caption{LiDAR-based panoptic segmentation results on the validation set of nuScenes. All results in [\%].}
    \vspace{-0.6cm}
    \begin{center}
    \small{
        \begin{tabular}{l|cccc|ccc|ccc|c}
            \Xhline{1pt}
            Method       & \PQ  & \PQda & \RQ  & \SQ  & \PQth & \RQth & \SQth & \PQst & \RQst & \SQst & \miou \\
            \hline\hline
            Cylinder3D \cite{zhou2020cylinder3d} +
            PointPillars \cite{lang2019pointpillars} & 36.0          & 44.5          & 43.0          & 83.3          & 23.3          & 27.0          & 83.7          & 57.2          & 69.6          & 82.7                    & 52.3          \\
            Cylinder3D \cite{zhou2020cylinder3d} +
            SECOND \cite{yan2018second}              & 40.1          & 48.4          & 47.3          & \textbf{84.2} & 29.0          & 33.6          & \textbf{84.4} & 58.5          & 70.1          & 83.7                    & 58.5          \\
            \hline
            \nickname{}                              & \textbf{42.5} & \textbf{51.0} & \textbf{50.3} & 83.6          & \textbf{32.5} & \textbf{38.3} & 83.1          & \textbf{59.2} & \textbf{70.3} & \textbf{84.4} & \textbf{70.7} \\
            \Xhline{1pt}
        \end{tabular}
    }
    \end{center}
    \label{tab:nus_val}
    \vspace{-0.8cm}
\end{table*}

\subsection{Evaluation Comparisons on nuScenes}

\noindent\textbf{Comparison Methods.}
Similarly, two strong semantic segmentation + detection baselines are provided for comparison on nuScenes.
The semantic segmentation method is Cylinder3D \cite{zhou2020cylinder3d} and the detection
methods are SECOND \cite{yan2018second} and PointPillars \cite{lang2019pointpillars}.
For fair comparison, the detection networks are trained using single frames on nuScenes.
The point-wise semantic predictions and predicted bounding boxes are merged in the following steps.
First all points inside each bounding box are assigned a unique instance IDs across the frame.
Then to unify the semantic predictions inside each instance, we assign the class labels of bounding
boxes predicted by the detection network to corresponding instances.

\noindent\textbf{Evaluation Results.}
%
As shown in Table \ref{tab:nus_val}, our \nickname{} outperforms the best baseline method in most metrics.
Especially, we surpass the best baseline method by 2.4\% in \PQ{} and 3.5\% in \PQth{}.
Unlike SemanticKITTI, nuScenes is featured as extremely sparse point clouds in single frames which
adds even more difficulties to panoptic segmentation.
The results validate the generalizability and the effectiveness of our \nickname{}.

\subsection{Further Analysis}
\noindent\textbf{Robust to Parameter Settings.}
As shown in Table \ref{tab:ab_bc}, six sets of bandwidth candidates are set for independent
training and the corresponding results are reported.
The stable results show that \nickname{} is robust to different parameter settings as long
as the picked bandwidth candidates are comparable to the instance sizes.
Unlike previous heuristic clustering algorithms that require massive parameter adjustment,
\nickname{} can automatically adjust to different instance sizes and point distributions and
remains stable clustering quality.
Further analyses on the iteration number settings are shown in supplementary materials.
\begin{table}[h]
    \caption{Results of different bandwidth candidates settings. All results in [\%].}
    \vspace{-0.6cm}
    \begin{center}
    \small{
        \begin{tabular}{l|cccc|c}
            \Xhline{1pt}
            \makecell{Bandwidth\\ Candidates (m)} & \PQ & \PQda & \RQ & \SQ & \miou \\
            \hline\hline
            0.2, 1.1, 2.0 & 57.4          & 63.0          & 67.7          & 77.4          & \textbf{63.7} \\
            0.2, 1.3, 2.4 & 57.5          & 63.1          & 67.7          & \textbf{77.6} & 63.5          \\
            0.2, 1.5, 2.8 & 57.6          & 63.2          & 67.8          & \textbf{77.6} & \textbf{63.7} \\
            0.2, 1.7, 3.2 & \textbf{57.7} & \textbf{63.4} & \textbf{68.0} & \textbf{77.6} & 63.5          \\
            0.2, 1.9, 3.6 & \textbf{57.7} & 63.3          & 67.9          & \textbf{77.6} & 63.4          \\
            0.2, 2.1, 4.0 & 57.4          & 63.1          & 67.7          & 77.5          & 63.3          \\
            \Xhline{1pt}
        \end{tabular}
    }
    \end{center}
    \label{tab:ab_bc}
    \vspace{-0.3cm}
\end{table}

\begin{figure}[ht]
    \vspace{-0.25cm}
    \begin{center}
        \includegraphics[width=1.0\linewidth]{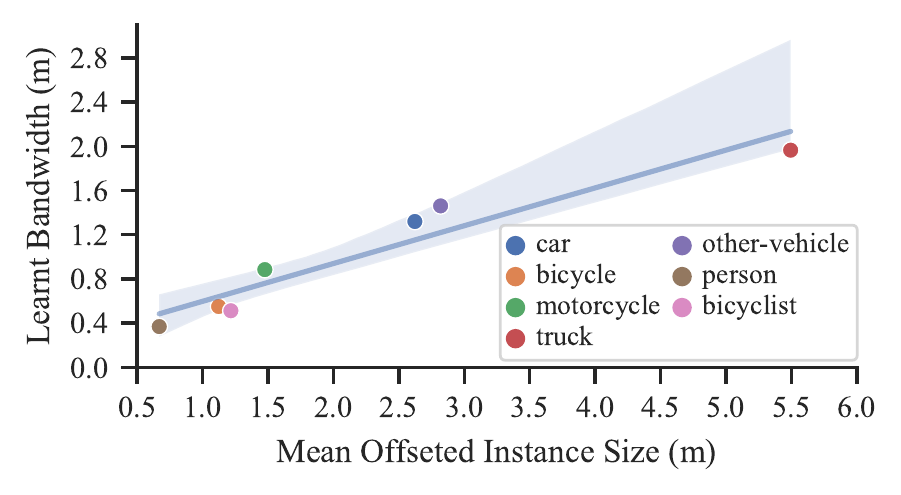}
    \end{center}
    \vspace{-0.8cm}
    \caption{\textbf{Proportional Relationship Between Sizes and the Learned Bandwidths.} The $x$-axis represents the class-wise average size of regressed centers of instances while the $y$-axis stands for the average learned bandwidth of different \things{} classes.}
    \label{fig:04_04_class_bandwidth}
    \vspace{-0.2cm}
\end{figure}

\noindent\textbf{Interpretable Learned Bandwidths.}
By averaging the bandwidth candidates weighted by the learned weights, the learned bandwidths for every
points could be approximated.
The average learned bandwidths of different classes are shown in Fig.
\ref{fig:04_04_class_bandwidth}.
The average learned bandwidths are roughly proportional to the instance sizes of corresponding classes, which is
consistent with the expectation that dynamic shifting can dynamically adjust to different instance sizes.
\begin{figure}[ht]
    \begin{center}
        \includegraphics[width=1.0\linewidth]{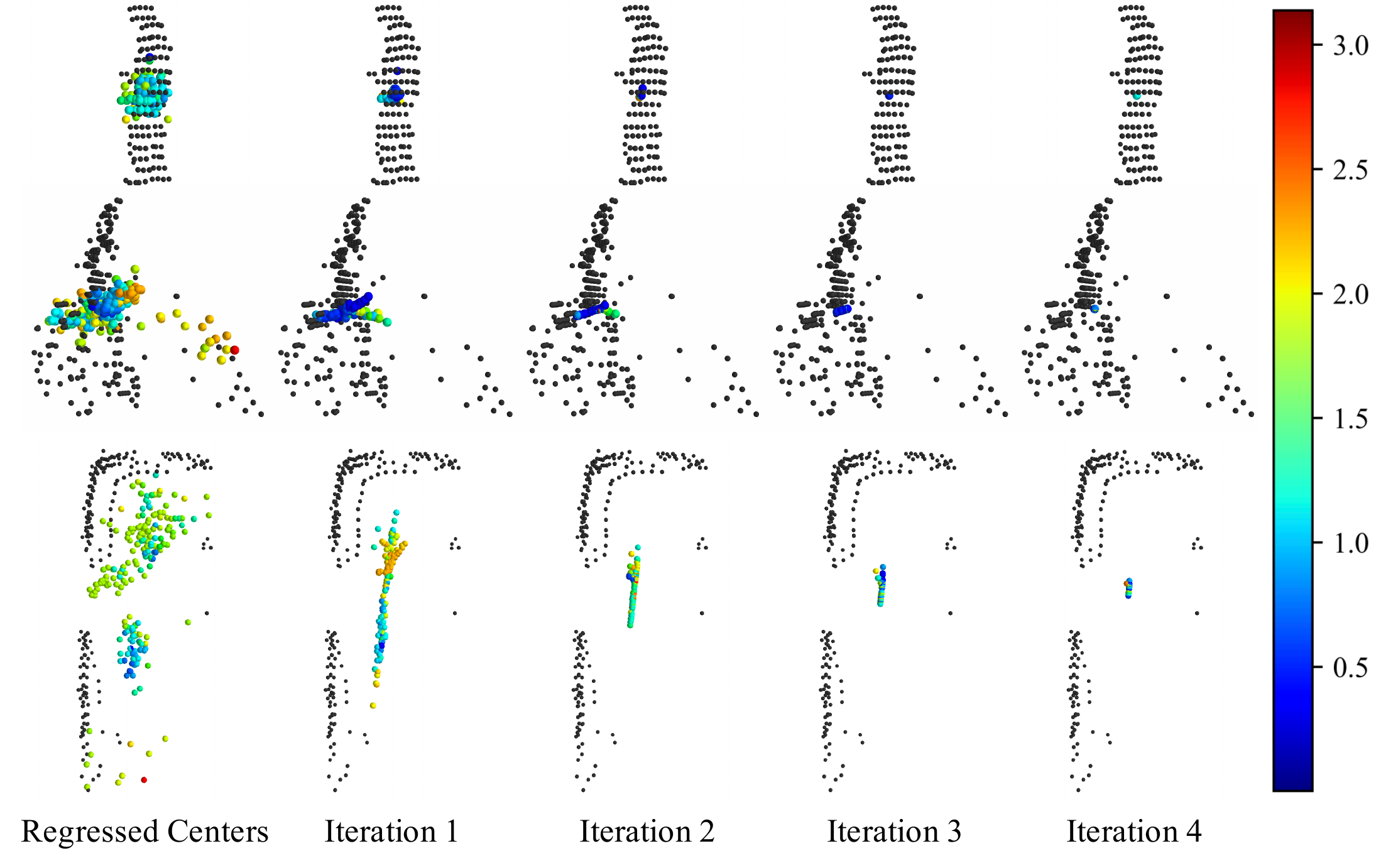}
    \end{center}
    \vspace{-0.6cm}
    \caption{\textbf{Visualization of Dynamic Shifting Iterations.} The black points are the original LiDAR point clouds of instances. The colored points are seeding points. From left to right, with the iteration number increases, the seeding points converge to cluster centers.}
    \label{fig:04_04_diffpart_bandwidth}
\end{figure}


\noindent\textbf{Visualization of Dynamic Shifting Iterations.}
As visualized in Fig. \ref{fig:04_04_diffpart_bandwidth}, the black points are
the original point clouds of different instances including person, bicyclist and car.
The seeding points are colored in spectral colors where the redder points represents higher learned bandwidth
and bluer points represents lower learned bandwidth.
The seeding points farther away from the instance centers tend to learn higher
bandwidths in order to quickly converge.
While the well-learned regressed points tend to have lower bandwidths to maintain their positions.
After four iterations, the seeding points have converged around the instance centers.
More visualization results could be found in supplementary materials.

\section{Conclusion}
With the goal of providing holistic perception for autonomous driving, we are one of the first to address the
task of LiDAR-based panoptic segmentation.
In order to tackle the challenge brought by the non-uniform distributions of LiDAR point clouds, we propose the
novel \nickname{} which is specifically designed for effective panoptic segmentation of LiDAR point clouds.
Our \nickname{} adopts strong baseline design which provides strong support for the consensus-driven fusion module
and the novel dynamic shifting module.
The novel dynamic shifting module adaptively shifts regressed centers of instances with different density and varying sizes.
The consensus-driven fusion efficiently unifies semantic and instance results into panoptic segmentation results.
The \nickname{} outperforms all strong baselines on both SemanticKITTI and nuScenes.
Further analyses show the robustness of the dynamic shifting module and the interpretability of the learned bandwidths.



{\small
    \bibliographystyle{ieee_fullname}
    \bibliography{egbib}

\begin{thebibliography}{10}\itemsep=-1pt

\bibitem{armeni20163d}
Iro Armeni, Ozan Sener, Amir~R Zamir, Helen Jiang, Ioannis Brilakis, Martin
  Fischer, and Silvio Savarese.
\newblock 3d semantic parsing of large-scale indoor spaces.
\newblock In {\em Proceedings of the IEEE Conference on Computer Vision and
  Pattern Recognition}, pages 1534--1543, 2016.

\bibitem{behley2019semantickitti}
Jens Behley, Martin Garbade, Andres Milioto, Jan Quenzel, Sven Behnke, Cyrill
  Stachniss, and Jurgen Gall.
\newblock Semantickitti: A dataset for semantic scene understanding of lidar
  sequences.
\newblock In {\em Proceedings of the IEEE International Conference on Computer
  Vision}, pages 9297--9307, 2019.

\bibitem{behley2020benchmark}
Jens Behley, Andres Milioto, and Cyrill Stachniss.
\newblock A benchmark for lidar-based panoptic segmentation based on kitti.
\newblock {\em arXiv preprint arXiv:2003.02371}, 2020.

\bibitem{berman2018lovasz}
Maxim Berman, Amal Rannen~Triki, and Matthew~B Blaschko.
\newblock The lov{\'a}sz-softmax loss: A tractable surrogate for the
  optimization of the intersection-over-union measure in neural networks.
\newblock In {\em Proceedings of the IEEE Conference on Computer Vision and
  Pattern Recognition}, pages 4413--4421, 2018.

\bibitem{nuscenes2019}
Holger Caesar, Varun Bankiti, Alex~H. Lang, Sourabh Vora, Venice~Erin Liong,
  Qiang Xu, Anush Krishnan, Yu Pan, Giancarlo Baldan, and Oscar Beijbom.
\newblock nuscenes: A multimodal dataset for autonomous driving.
\newblock {\em arXiv preprint arXiv:1903.11027}, 2019.

\bibitem{campello2013density}
Ricardo~JGB Campello, Davoud Moulavi, and J{\"o}rg Sander.
\newblock Density-based clustering based on hierarchical density estimates.
\newblock In {\em Pacific-Asia conference on knowledge discovery and data
  mining}, pages 160--172. Springer, 2013.

\bibitem{choy20194d}
Christopher Choy, JunYoung Gwak, and Silvio Savarese.
\newblock 4d spatio-temporal convnets: Minkowski convolutional neural networks.
\newblock In {\em Proceedings of the IEEE Conference on Computer Vision and
  Pattern Recognition}, pages 3075--3084, 2019.

\bibitem{comaniciu2002mean}
Dorin Comaniciu and Peter Meer.
\newblock Mean shift: A robust approach toward feature space analysis.
\newblock {\em IEEE Transactions on pattern analysis and machine intelligence},
  24(5):603--619, 2002.

\bibitem{dai2017scannet}
Angela Dai, Angel~X Chang, Manolis Savva, Maciej Halber, Thomas Funkhouser, and
  Matthias Nie{\ss}ner.
\newblock Scannet: Richly-annotated 3d reconstructions of indoor scenes.
\newblock In {\em Proceedings of the IEEE Conference on Computer Vision and
  Pattern Recognition}, pages 5828--5839, 2017.

\bibitem{engelmann20203d}
Francis Engelmann, Martin Bokeloh, Alireza Fathi, Bastian Leibe, and Matthias
  Nie{\ss}ner.
\newblock 3d-mpa: Multi-proposal aggregation for 3d semantic instance
  segmentation.
\newblock In {\em Proceedings of the IEEE/CVF Conference on Computer Vision and
  Pattern Recognition}, pages 9031--9040, 2020.

\bibitem{ester1996density}
Martin Ester, Hans-Peter Kriegel, J{\"o}rg Sander, Xiaowei Xu, et~al.
\newblock A density-based algorithm for discovering clusters in large spatial
  databases with noise.
\newblock In {\em Kdd}, volume~96, pages 226--231, 1996.

\bibitem{han2020occuseg}
Lei Han, Tian Zheng, Lan Xu, and Lu Fang.
\newblock Occuseg: Occupancy-aware 3d instance segmentation.
\newblock In {\em Proceedings of the IEEE/CVF Conference on Computer Vision and
  Pattern Recognition}, pages 2940--2949, 2020.

\bibitem{hu2020randla}
Qingyong Hu, Bo Yang, Linhai Xie, Stefano Rosa, Yulan Guo, Zhihua Wang, Niki
  Trigoni, and Andrew Markham.
\newblock Randla-net: Efficient semantic segmentation of large-scale point
  clouds.
\newblock In {\em Proceedings of the IEEE/CVF Conference on Computer Vision and
  Pattern Recognition}, pages 11108--11117, 2020.

\bibitem{jiang2020pointgroup}
Li Jiang, Hengshuang Zhao, Shaoshuai Shi, Shu Liu, Chi-Wing Fu, and Jiaya Jia.
\newblock Pointgroup: Dual-set point grouping for 3d instance segmentation.
\newblock In {\em Proceedings of the IEEE/CVF Conference on Computer Vision and
  Pattern Recognition}, pages 4867--4876, 2020.

\bibitem{kirillov2019panoptic}
Alexander Kirillov, Kaiming He, Ross Girshick, Carsten Rother, and Piotr
  Doll{\'a}r.
\newblock Panoptic segmentation.
\newblock In {\em Proceedings of the IEEE conference on computer vision and
  pattern recognition}, pages 9404--9413, 2019.

\bibitem{kong2018recurrent}
Shu Kong and Charless~C Fowlkes.
\newblock Recurrent pixel embedding for instance grouping.
\newblock In {\em Proceedings of the IEEE Conference on Computer Vision and
  Pattern Recognition}, pages 9018--9028, 2018.

\bibitem{lahoud20193d}
Jean Lahoud, Bernard Ghanem, Marc Pollefeys, and Martin~R Oswald.
\newblock 3d instance segmentation via multi-task metric learning.
\newblock In {\em Proceedings of the IEEE International Conference on Computer
  Vision}, pages 9256--9266, 2019.

\bibitem{lang2019pointpillars}
Alex~H Lang, Sourabh Vora, Holger Caesar, Lubing Zhou, Jiong Yang, and Oscar
  Beijbom.
\newblock Pointpillars: Fast encoders for object detection from point clouds.
\newblock In {\em Proceedings of the IEEE Conference on Computer Vision and
  Pattern Recognition}, pages 12697--12705, 2019.

\bibitem{milioto2020iros}
A. Milioto, J. Behley, C. McCool, and C. Stachniss.
\newblock Lidar panoptic segmentation for autonomous driving.
\newblock In {\em IROS}, 2020.

\bibitem{milioto2019rangenet++}
Andres Milioto, Ignacio Vizzo, Jens Behley, and Cyrill Stachniss.
\newblock Rangenet++: Fast and accurate lidar semantic segmentation.
\newblock In {\em 2019 IEEE/RSJ International Conference on Intelligent Robots
  and Systems (IROS)}, pages 4213--4220. IEEE, 2019.

\bibitem{pham2019jsis3d}
Quang-Hieu Pham, Thanh Nguyen, Binh-Son Hua, Gemma Roig, and Sai-Kit Yeung.
\newblock Jsis3d: joint semantic-instance segmentation of 3d point clouds with
  multi-task pointwise networks and multi-value conditional random fields.
\newblock In {\em Proceedings of the IEEE Conference on Computer Vision and
  Pattern Recognition}, pages 8827--8836, 2019.

\bibitem{qi2017pointnet}
Charles~R Qi, Hao Su, Kaichun Mo, and Leonidas~J Guibas.
\newblock Pointnet: Deep learning on point sets for 3d classification and
  segmentation.
\newblock In {\em Proceedings of the IEEE conference on computer vision and
  pattern recognition}, pages 652--660, 2017.

\bibitem{qi2017pointnet++}
Charles~Ruizhongtai Qi, Li Yi, Hao Su, and Leonidas~J Guibas.
\newblock Pointnet++: Deep hierarchical feature learning on point sets in a
  metric space.
\newblock In {\em Advances in neural information processing systems}, pages
  5099--5108, 2017.

\bibitem{shi2020pv}
Shaoshuai Shi, Chaoxu Guo, Li Jiang, Zhe Wang, Jianping Shi, Xiaogang Wang, and
  Hongsheng Li.
\newblock Pv-rcnn: Point-voxel feature set abstraction for 3d object detection.
\newblock In {\em Proceedings of the IEEE/CVF Conference on Computer Vision and
  Pattern Recognition}, pages 10529--10538, 2020.

\bibitem{thomas2019kpconv}
Hugues Thomas, Charles~R Qi, Jean-Emmanuel Deschaud, Beatriz Marcotegui,
  Fran{\c{c}}ois Goulette, and Leonidas~J Guibas.
\newblock Kpconv: Flexible and deformable convolution for point clouds.
\newblock In {\em Proceedings of the IEEE International Conference on Computer
  Vision}, pages 6411--6420, 2019.

\bibitem{wang2018sgpn}
Weiyue Wang, Ronald Yu, Qiangui Huang, and Ulrich Neumann.
\newblock Sgpn: Similarity group proposal network for 3d point cloud instance
  segmentation.
\newblock In {\em Proceedings of the IEEE Conference on Computer Vision and
  Pattern Recognition}, pages 2569--2578, 2018.

\bibitem{wang2019associatively}
Xinlong Wang, Shu Liu, Xiaoyong Shen, Chunhua Shen, and Jiaya Jia.
\newblock Associatively segmenting instances and semantics in point clouds.
\newblock In {\em Proceedings of the IEEE Conference on Computer Vision and
  Pattern Recognition}, pages 4096--4105, 2019.

\bibitem{wang2019dynamic}
Yue Wang, Yongbin Sun, Ziwei Liu, Sanjay~E Sarma, Michael~M Bronstein, and
  Justin~M Solomon.
\newblock Dynamic graph cnn for learning on point clouds.
\newblock {\em Acm Transactions On Graphics (tog)}, 38(5):1--12, 2019.

\bibitem{Wong2019IdentifyingUI}
Kelvin Wong, Shenlong Wang, Mengye Ren, Ming Liang, and Raquel Urtasun.
\newblock Identifying unknown instances for autonomous driving.
\newblock In {\em The Conference on Robot Learning ({CORL})}, 2019.

\bibitem{wu2018squeezeseg}
Bichen Wu, Alvin Wan, Xiangyu Yue, and Kurt Keutzer.
\newblock Squeezeseg: Convolutional neural nets with recurrent crf for
  real-time road-object segmentation from 3d lidar point cloud.
\newblock In {\em 2018 IEEE International Conference on Robotics and Automation
  (ICRA)}, pages 1887--1893. IEEE, 2018.

\bibitem{wu2019pointconv}
Wenxuan Wu, Zhongang Qi, and Li Fuxin.
\newblock Pointconv: Deep convolutional networks on 3d point clouds.
\newblock In {\em Proceedings of the IEEE Conference on Computer Vision and
  Pattern Recognition}, pages 9621--9630, 2019.

\bibitem{yan2018second}
Yan Yan, Yuxing Mao, and Bo Li.
\newblock Second: Sparsely embedded convolutional detection.
\newblock {\em Sensors}, 18(10):3337, 2018.

\bibitem{zhang2020spatial}
Dongsu Zhang, Junha Chun, Sang~Kyun Cha, and Young~Min Kim.
\newblock Spatial semantic embedding network: Fast 3d instance segmentation
  with deep metric learning.
\newblock {\em arXiv preprint arXiv:2007.03169}, 2020.

\bibitem{2020LiDARSeg}
Feihu Zhang, Chenye Guan, Jin Fang, Song Bai, Ruigang Yang, Philip Torr, and
  Victor Prisacariu.
\newblock Instance segmentation of lidar point clouds.
\newblock {\em ICRA, Cited by}, 4(1), 2020.

\bibitem{zhang2020polarnet}
Yang Zhang, Zixiang Zhou, Philip David, Xiangyu Yue, Zerong Xi, Boqing Gong,
  and Hassan Foroosh.
\newblock Polarnet: An improved grid representation for online lidar point
  clouds semantic segmentation.
\newblock In {\em Proceedings of the IEEE/CVF Conference on Computer Vision and
  Pattern Recognition}, pages 9601--9610, 2020.

\bibitem{zhao2020jsnet}
Lin Zhao and Wenbing Tao.
\newblock Jsnet: Joint instance and semantic segmentation of 3d point clouds.
\newblock In {\em AAAI}, pages 12951--12958, 2020.

\bibitem{zhou2020cylinder3d}
Hui Zhou, Xinge Zhu, Xiao Song, Yuexin Ma, Zhe Wang, Hongsheng Li, and Dahua
  Lin.
\newblock Cylinder3d: An effective 3d framework for driving-scene lidar
  semantic segmentation.
\newblock {\em arXiv preprint arXiv:2008.01550}, 2020.

\end{thebibliography}
}

\clearpage

\setcounter{section}{0}

\begin{center}
 {\large \textbf{Appendix}}
\end{center}

In this appendix, we provide the following sections for a better understanding of the main paper.
Firstly, the details of the heuristic clustering algorithms mentioned in the main paper are provided (Sec. 1).
Then, we provide analyses of the differentiability of the dynamic shifting to give some insights into our design (Sec. 2).
We further analyze the number of iterations in the dynamic shifting module (Sec. 3).
Moreover, we report the implementation details of \nickname{} for the reproducibility (Sec. 4).
Last but not least, the per-class results are reported (Sec. 5), and more visualization examples are displayed (Sec. 6).

\section{Details of Heuristic Clustering Algorithms} \label{sec:1}
\noindent\textbf{Breadth First Search (BFS).}
BFS is simple but effective for indoor point clouds.
For the points to be clustered, BFS first constructs a graph where edges connect point
pairs that are closer than a given radius.
Then each connected sub-graph is considered a cluster.
BFS has shown that it is capable of performing high-quality clustering for
indoor point clouds \cite{jiang2020pointgroup} which are dense, even and complete.
However, it does not apply to LiDAR point clouds.
As discussed in the main paper, large density difference within and between clusters means that the fixed radius can
not properly adapt to different clusters.
Therefore, it is not a good idea to use BFS as the clustering algorithm for autonomous driving scenes.

\noindent\textbf{DBSCAN \cite{ester1996density} and HDBSCAN \cite{campello2013density}.}
Both DBSCAN and HDBSCAN are density-based clustering algorithms which make them perform badly on LiDAR point clouds.
Similar to BFS, DBSCAN also constructs a graph based on the mutual distances of the points.
Although the new concept of \textit{core point} is introduced to filter out noise points, the problem brought by the fixed radius
also occurs in this algorithm.
Moreover, the mechanism of noise points recognition also brings problems.
For any points to be clustered, if there exists less than a certain number of points within a certain radius, the point is labeled as a noise point.
However, the number and densities of points inside instances vary greatly, which makes it hard to determine the line between instances with little
points and noise points.

HDBSCAN has a more complex rule of constructing graphs and is claimed to be more robust to density changing than
DBSCAN.
The \textit{mutual reachable distance} replaced euclidean distance as the indicators of graph construction, which makes it more robust to
density changes.
Moreover, with the help of the cluster hierarchy, DBSCAN can automatically adapt to data with different distributions.
However, by introducing the concept of \textit{mutual reachable distance}, HDBSCAN intuitively assumes that the points with
lower density are more likely to be \textit{seas} (noise points) that separate \textit{lands} (valid clusters), which
is not the case in LiDAR point clouds where low-density point clouds can also be valid instances that are far away
from the LiDAR sensor.
Thus DBSCAN and HDBSCAN can not provide high-quality clustering results.

\noindent\textbf{Mean Shift \cite{comaniciu2002mean}.}
Mean Shift performs clustering in a very different way than the above three algorithms.
Firstly, seeding points are sampled from the points to be clustered.
Then, seeding points are iteratively shifted towards cluster centers in order
to obtain centers of all clusters.
The positions where seeding points are shifted to is calculated by applying the kernel function
on corresponding points.
In our implementation, we use the flat kernel which takes the mean of all points within a query ball
as the result.
The radius of the ball is denoted \textit{bandwidth}.
After several iterations, all the shifted points have converged and the cluster centers are extracted from the converged points.
All the points to be clustered are assigned to the nearest cluster centers, which produces the final instance IDs.
The advantage of Mean Shift is that the kernel function is not sensitive to density changes and robust to noise
points, which makes it more suitable than density-based clustering algorithms.

However, Mean Shift is not perfect.
The choice of parameters of the kernel function, which is the \textit{bandwidth} in this case, is not trivial.
The bandwidth controls the range that the kernel function is applied on.
Small bandwidth would mislead the regressed centers of a single instance shifting to several different cluster
centers and cause over-segmentation.
On the contrary, large bandwidth would mislead regressed centers of neighboring instances shifting to one cluster
center and result in under-segmentation.
The performance of the classes with relatively small size drops with the bandwidth increasing and vice versa.
Therefore, the fixed bandwidth can not handle large and small instances simultaneously.
Besides, assigning each point to the cluster centers is not reasonable for that nearby instances may have
different sizes which would lead to a different degree of dispersion of regressed centers.
For example, edge points of a large instance may be farther to its center than that of nearby instances.
Although Mean Shift is not as bad as density-based clustering algorithms, there remains a lot of room for improvement.

\section{Gradient Calculation of Dynamic Shifting} \label{sec:2}
In this chapter, we are going to show the gradients of one dynamic shifting iteration and that directly regressing bandwidth
with flat kernel is not differentiable.
The forward pass of dynamic shifting can be broken down to the following steps:
\begin{align}
    K_j &= (XX^{T} \leq \delta_{j}) \label{eq:k} \\
    D_j &= K_j \mathbf{1} \label{eq:d} \\
    S_j &= D_j^{-1} K_j X \label{eq:s} \\
    w_j &= f(F, p_j) \\
    W_j &= w_j \mathbf{1}^{1\times 3} \\
    Y_j &= W_j \odot S_j \\
    Y   &= \sum_{j=1}^{l} Y_j \text{,}
\end{align}
where $X$ represents the seeding points, $f$ is the combination of Softmax and MLP, $p_j$ is the parameter of $f$, $F$
is the backbone features, $Y$ is the shifted seeding points, and $\odot$ denotes element-wise product.
It is worth noting that $S_j$, which is defined by equation \ref{eq:k}, \ref{eq:d} and \ref{eq:s}, is constant and therefore
does not have gradients.
Assuming $c$ is the loss, backpropagation gradients are calculated as follows.
\begin{align}
    \frac{\partial c}{\partial F}   &= \frac{\partial f(F, p_j)}{\partial F} \frac{\partial c}{\partial w_j} \\
    \frac{\partial c}{\partial p_j} &= \frac{\partial f(F, p_j)}{\partial p_j} \frac{\partial c}{\partial w_j} \\
    \frac{\partial c}{\partial w_j} &= \frac{\partial c}{\partial W_j} \mathbf{1}^{3\times 1}\\
    \frac{\partial c}{\partial W_j} &= S_j \odot \frac{\partial c}{\partial Y_j} \\
    \frac{\partial c}{\partial Y_j} &= \frac{\partial c}{\partial Y}
\end{align}

However, if the bandwidth $\delta$ is learnable, then equation \ref{eq:k} will turn to:
\begin{equation}
    K = (XX^{T} \leq \delta) \text{,}
\end{equation}
which unfortunately is not differentiable. Therefore, in our ablation study, in order to further demonstrate that direct regression
is not a good strategy, we adopt the Gaussian kernel which is formally defined as:
\begin{equation}
    K = exp(-\frac{XX^{T}}{2\delta^{2}}) \text{,}
\end{equation}
where $\delta$ is the learnable bandwidth. With the Gaussian kernel, the direct regression version of dynamic shifting
is now differentiable.

\section{Further Analyses of the Number of Iterations} \label{sec:3}
\noindent\textbf{Number of Iterations Settings.}
In the dynamic shifting module, other than bandwidth candidates, the hyper-parameter to tune is the number of iterations, which is
essential for the final clustering quality because too few iterations would cause insufficient
convergence while too many iterations would add unnecessary time and space complexity.
As shown in Table \ref{tab:ab_iter}, we experiment on several different iteration number settings.
The best result is achieved when the number of iterations is set to 4 which is the counterpoint of
sufficient convergence and efficiency.
\begin{table}[h]
    \caption{Experiments on the number of iteration. All results in [\%].}
    \vspace{-0.6cm}
    \begin{center}
    \small{
        \begin{tabular}{c|cccc|c}
            \Xhline{1pt}
            \makecell{Number of\\ Iteration} & \PQ & \PQda & \RQ & \SQ & \miou \\
            \hline\hline
            1 & 57.0          & 62.6          & 67.4          & 77.3          & 63.5          \\
            2 & 57.6          & 63.1          & 67.8          & 77.5          & \textbf{63.6} \\
            3 & \textbf{57.7} & 63.3          & \textbf{68.0} & \textbf{77.6} & 63.4          \\
            4 & \textbf{57.7} & \textbf{63.4} & \textbf{68.0} & \textbf{77.6} & 63.5          \\
            5 & 57.5          & 63.2          & 67.7          & \textbf{77.6} & 63.4          \\
            \Xhline{1pt}
        \end{tabular}
    }
    \end{center}
    \label{tab:ab_iter}
\end{table}

\begin{figure}[ht]
    \begin{center}
        \includegraphics[width=1.0\linewidth]{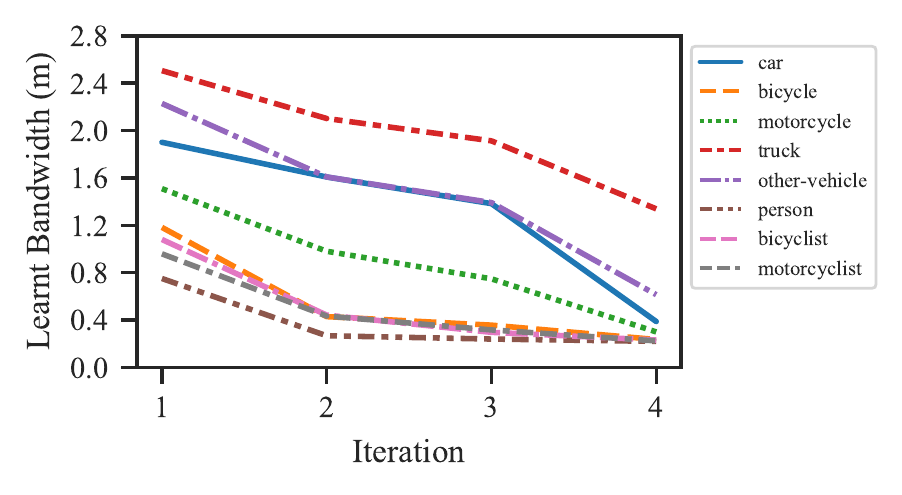}
    \end{center}
    \vspace{-0.8cm}
    \caption{\textbf{Relationship Between Iterations and the Learned Bandwidths.} With number of iteration increases, the learned bandwidth decreases. At the 4th iteration, the learned bandwidths of most classes drop near the lower limit.}
    \label{fig:04_04_iteration_bandwidth}
    \vspace{-0.3cm}
\end{figure}

\noindent\textbf{Learned Bandwidths of Different Iterations.}
The average learned bandwidths of different iterations are shown in Fig. \ref{fig:04_04_iteration_bandwidth}.
As expected, as the iteration rounds grow, points of the same instance gather tighter which usually require
smaller bandwidths.
After four iterations, learned bandwidths of most classes have dropped to 0.2, which is the lowest they
can get, meaning that four iterations are enough for \things{} points to converge to cluster centers, which
further validates the conclusion made in the last paragraph.

\section{Implementation Details} \label{sec:4}
\noindent\textbf{Backbone.}
For both datasets, each input point is represented as a $4$ dimension vector including XYZ coordinates and the intensity.
The backbone voxelizes a single frame to $480\times 360\times 32$ voxels under the cylindrical coordinate
system.
For that we should not use the information of bounding boxes in this segmentation task, the ground truth
center of each instance is approximated by the center of its tight box that parallel to axes which makes a better
approximation than the mass centers of the incomplete point clouds.
The bandwidth of the Mean Shift used in our backbone method is set to $1.2$.
Adam solver is utilized to optimize the network.
The minimum number of points in a valid instance is set to 50 for SemanticKITTI and 5 for nuScenes.

\noindent\textbf{Dynamic Shifting Module.}
The number of the FPS downsampled points in the dynamic shifting module is set to $10000$.
The final heuristic clustering algorithm used in the dynamic shifting module is Mean Shift with $0.65$ bandwidth for
SemanticKITTI and BFS with $1.2$ radius for nuScenes.
Bandwidth candidates are set to $0.2$, $1.7$ and $3.2$ for both datasets.
The number of Iterations is set to $4$ for both datasets.

\begin{table*}
    \caption{Detailed per-class \PQ{} results on the test set of SemanticKITTI.}
    \vspace{-0.6cm}
    \begin{center}
    \tabcolsep=0.13cm
    \scriptsize{
        \begin{tabular}{l|c|ccccccccccccccccccc}
            \Xhline{1pt}
                           
            Method & \PQ & \car  & \truc & \bcle & \mcle & \oveh & \pers & \bcli & \mcli & \road & \side & \park & \ogro & \buil & \vege & \trun & \terr & \fenc & \pole & \traf \\
            \hline\hline
            KPConv\cite{thomas2019kpconv} +
            PointPillars\cite{lang2019pointpillars} & 44.5 & 72.5 & 17.2 &  9.2 & 30.8 & 19.6 & 29.9 & 59.4 & 22.8 & 84.6 & 60.1 & 34.1 & \textbf{8.8} & 80.7 & 77.6 & 53.9 & 42.2 & 49.0 & 46.2 & 46.8 \\
            RangeNet++\cite{milioto2019rangenet++} +
            PointPillars\cite{lang2019pointpillars} & 37.1 & 66.9 &  6.7 &  3.1 & 16.2 &  8.8 & 14.6 & 31.8 & 13.5 & \textbf{90.6} & 63.2 & \textbf{41.3} &  6.7 & 79.2 & 71.2 & 34.6 & 37.4 & 38.2 & 32.8 & 47.4 \\
            KPConv\cite{thomas2019kpconv} +
            PV-RCNN\cite{shi2020pv}                 & 50.2 & 84.5 & 21.9 &  9.9 & 34.2 & 25.6 & 51.1 & 67.9 & 43.8 & 84.9 & \textbf{63.6} & 37.1 &  8.4 & 83.7 & 78.3 & 57.5 & 42.3 & \textbf{51.1} & 51.0 & 57.4 \\
            \hline
            PolarNet\_seg$^{\ast}$                  & 53.3 & 88.7 & \textbf{31.7} & 34.6 & \textbf{50.9} & \textbf{39.1} & 57.5 & 68.7 & 45.1 & 88.1 & 59.7 & 40.5 &  1.0 & \textbf{85.7} & 77.7 & 53.2 & 39.7 & 44.8 & 48.6 & 57.6 \\
            \hline
            Backbone with Fusion                    & 53.1 & 90.6 & 15.8 & 44.2 & 46.9 & 28.5 & 63.1 & 67.7 & 47.6 & 88.2 & 59.4 & 29.5 &  3.0 & 82.5 & 79.0 & 56.6 & 42.3 & 48.1 & 53.2 & 63.6\\
            \nickname{}                             & \textbf{55.9} & \textbf{91.2} & 28.8 & \textbf{45.4} & 47.2 & 34.6 & \textbf{63.6} & \textbf{71.1} & \textbf{58.5} & 89.1 & 61.2 & 32.3 &  4.0 & 83.2 & \textbf{79.6} & \textbf{58.3} & \textbf{43.4} & 50.0 & \textbf{55.2} & \textbf{65.3} \\
            \Xhline{1pt}
        \end{tabular}
    }
    \end{center}
    \label{tab:semkitti_test_pq}
    \vspace{-0.6cm}
\end{table*}

\begin{table*}
    \caption{Detailed per-class \RQ{} results on the test set of SemanticKITTI.}
    \vspace{-0.6cm}
    \begin{center}
    \tabcolsep=0.13cm
    \scriptsize{
        \begin{tabular}{l|c|ccccccccccccccccccc}
            \Xhline{1pt}
                           
            Method & \RQ & \car  & \truc & \bcle & \mcle & \oveh & \pers & \bcli & \mcli & \road & \side & \park & \ogro & \buil & \vege & \trun & \terr & \fenc & \pole & \traf \\
            \hline\hline
            KPConv\cite{thomas2019kpconv} +
            PV-RCNN\cite{shi2020pv}                 & 61.4 & 94.5 & 26.9 & 14.5 & 43.6 & 32.0 & 70.0 & 76.8 & 52.6 & 91.5 & \textbf{78.7} & 47.7 & \textbf{11.4} & 90.2 & 94.1 & 76.4 & 56.5 & \textbf{66.8} & 68.8 & 74.1 \\
            \hline
            PolarNet\_seg$^{\ast}$                  & 64.2 & 96.2 & \textbf{36.0} & 48.5 & \textbf{58.6} & \textbf{43.0} & 66.2 & 77.1 & 50.3 & 96.3 & 74.8 & \textbf{54.2} &  1.7 & \textbf{92.1} & 94.1 & 72.6 & 53.9 & 61.1 & 66.0 & 76.3 \\
            \hline
            Backbone with Fusion                    & 64.5 & 97.4 & 20.5 & 61.1 & 57.2 & 33.2 & 74.0 & 75.6 & 57.8 & 96.2 & 75.1 & 39.2 & 5.0 & 88.8 & 95.5 & 75.8 & 57.1 & 63.8 & 72.2 & 80.4 \\
            \nickname{}                             & \textbf{66.7} & \textbf{97.5} & 32.4 & \textbf{62.2} & 56.3 & 38.9 & \textbf{74.3} & \textbf{78.4} & \textbf{62.7} & \textbf{96.8} & 76.7 & 42.6 & 6.4 & 89.3 & \textbf{95.7} & \textbf{77.5} & \textbf{58.3} & 65.5 & \textbf{74.0} & \textbf{81.9} \\
            \Xhline{1pt}
        \end{tabular}
    }
    \end{center}
    \label{tab:semkitti_test_rq}
    \vspace{-0.6cm}
\end{table*}

\begin{table*}
    \caption{Detailed per-class \SQ{} results on the test set of SemanticKITTI.}
    \vspace{-0.6cm}
    \begin{center}
    \tabcolsep=0.13cm
    \scriptsize{
        \begin{tabular}{l|c|ccccccccccccccccccc}
            \Xhline{1pt}
                           
            Method & \SQ & \car  & \truc & \bcle & \mcle & \oveh & \pers & \bcli & \mcli & \road & \side & \park & \ogro & \buil & \vege & \trun & \terr & \fenc & \pole & \traf \\
            \hline\hline
            KPConv\cite{thomas2019kpconv} +
            PV-RCNN\cite{shi2020pv}                 & 80.0 & 89.4 & 81.3 & 68.1 & 78.4 & 79.8 & 73.0 & 88.4 & 83.3 & \textbf{92.9} & \textbf{80.8} & \textbf{77.8} & \textbf{73.4} & 92.8 & \textbf{83.2} & \textbf{75.2} & \textbf{75.0} & \textbf{76.5} & 74.2 & 77.5 \\
            \hline
            PolarNet\_seg$^{\ast}$                  & 81.1 & 92.2 & 88.0 & 71.4 & \textbf{86.9} & \textbf{90.8} & \textbf{87.0} & 89.1 & 89.8 & 91.4 & 79.9 & 74.8 & 55.4 & 93.1 & 82.6 & 73.3 & 73.8 & 73.2 & 73.6 & 75.5 \\
            \hline
            Backbone with Fusion                    & 80.3 & 93.0 & 77.0 & 72.3 & 81.9 & 85.8 & 85.3 & 89.5 & 82.5 & 91.8 & 79.1 & 75.1 & 60.2 & 92.9 & 82.7 & 74.7 & 74.0 & 75.5 & 73.6 & 79.1 \\
            \nickname{}                             & \textbf{82.3} & \textbf{93.6} & \textbf{88.9} & \textbf{73.0} & 83.8 & 89.0 & 85.6 & \textbf{90.7} & \textbf{93.3} & 92.0 & 79.8 & 75.8 & 61.4 & \textbf{93.2} & \textbf{83.2} & \textbf{75.2} & 74.4 & 76.3 & \textbf{74.5} & \textbf{79.7} \\
            \Xhline{1pt}
        \end{tabular}
    }
    \end{center}
    \label{tab:semkitti_test_sq}
    \vspace{-0.6cm}
\end{table*}

\begin{figure*}[t]
    \begin{center}
        \includegraphics[width=1.0\linewidth]{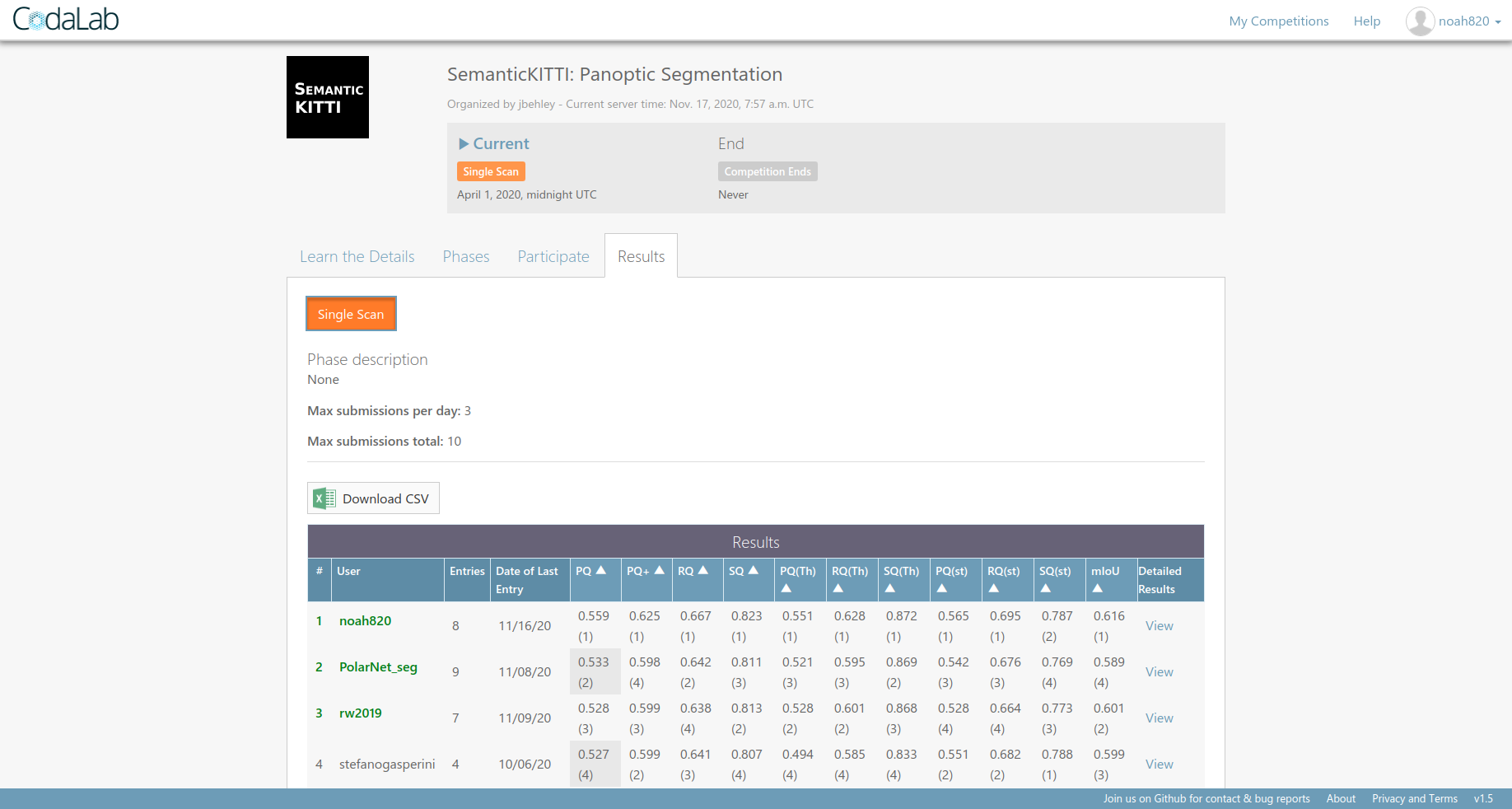}
    \end{center}
    \caption{Screenshot of the public leaderboard (\url{https://competitions.codalab.org/competitions/24025}) of SemanticKITTI at 2020-11-16. Our method achieves 1st place.}
    \label{fig:supp_colab}
\end{figure*}

\section{Per-class Evaluation Results} \label{sec:5}
Detailed per-class \PQ{}, \RQ{} and \SQ{} results are presented in table \ref{tab:semkitti_test_pq}, \ref{tab:semkitti_test_rq} and \ref{tab:semkitti_test_sq} respectively.
All the results are reported on the held-out test set of SemanticKITTI.
``$\ast$'' denotes the unpublished method which is in 2nd place on the public benchmark of SemanticKITTI (accessed on 2020-11-16).
Compared to semantic segmentation + detection baseline methods, our \nickname{} has huge advantages in \things{} classes in terms of \PQ{}, \RQ{} and \SQ{}.
With the help of the dynamic shifting module, our \nickname{} surpasses the backbone (with fusion module) in all classes in all three metrics, which demonstrates the effectiveness
of the novel dynamic shifting module.
Moreover, our \nickname{} shows superiority in most classes compared with 2nd place method ``Polarnet\_seg''.
Fig. \ref{fig:supp_colab} gives the screenshot of the public leaderboard of SemanticKITTI at 2020-11-16 and our \nickname{} achieves 1st place.

\section{More Visualization Results of the \nickname{}} \label{sec:6}
\begin{figure*}[t]
    \vspace{-0.6cm}
    \begin{center}
        \includegraphics[width=0.9\linewidth]{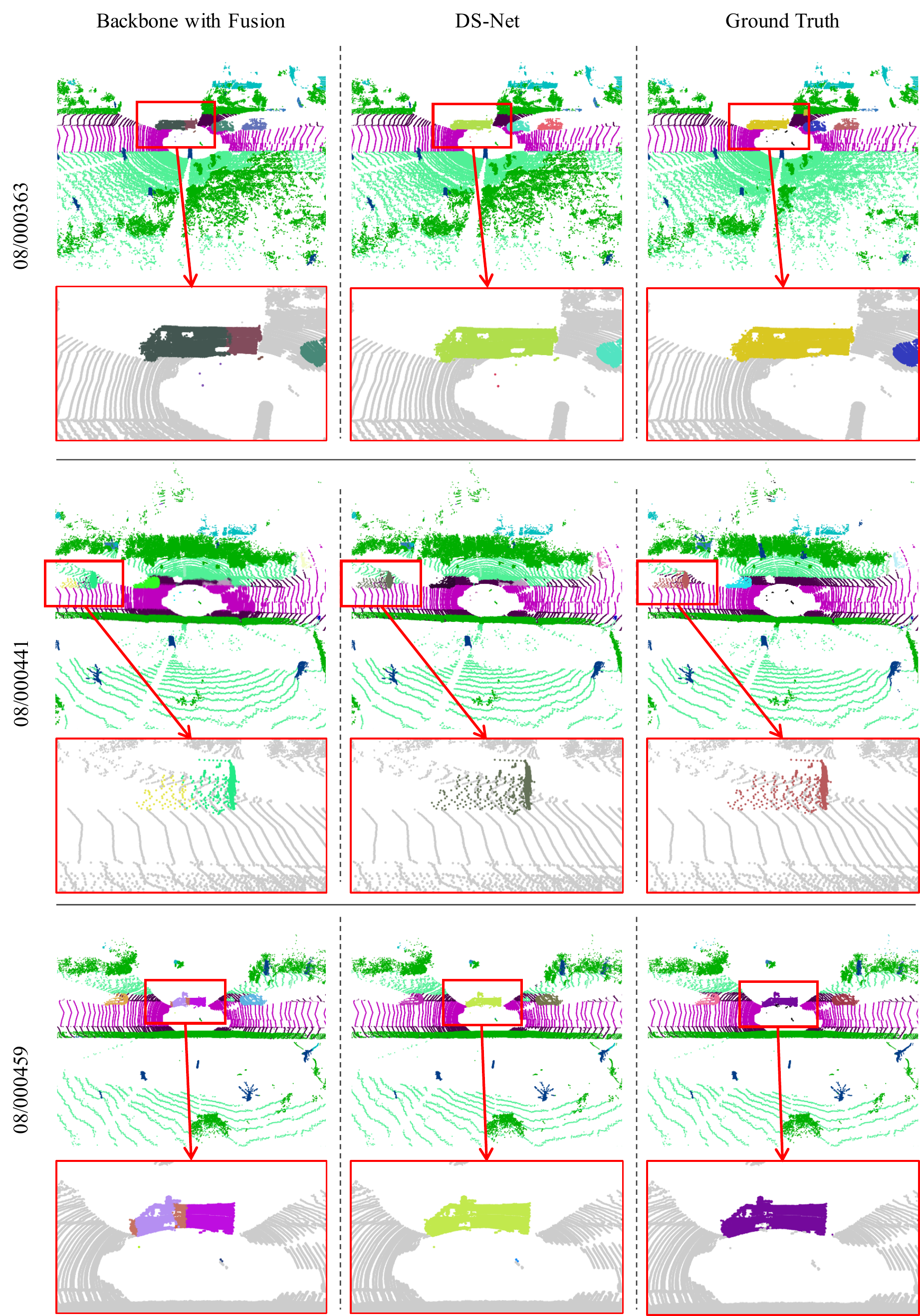}
    \end{center}
    \vspace{-0.6cm}
    \caption{\textbf{Qualitative Comparison of the Backbone and \nickname{} (1).}}
    \label{fig:vis_01}
\end{figure*}

\begin{figure*}[t]
    \vspace{-0.6cm}
    \begin{center}
        \includegraphics[width=0.9\linewidth]{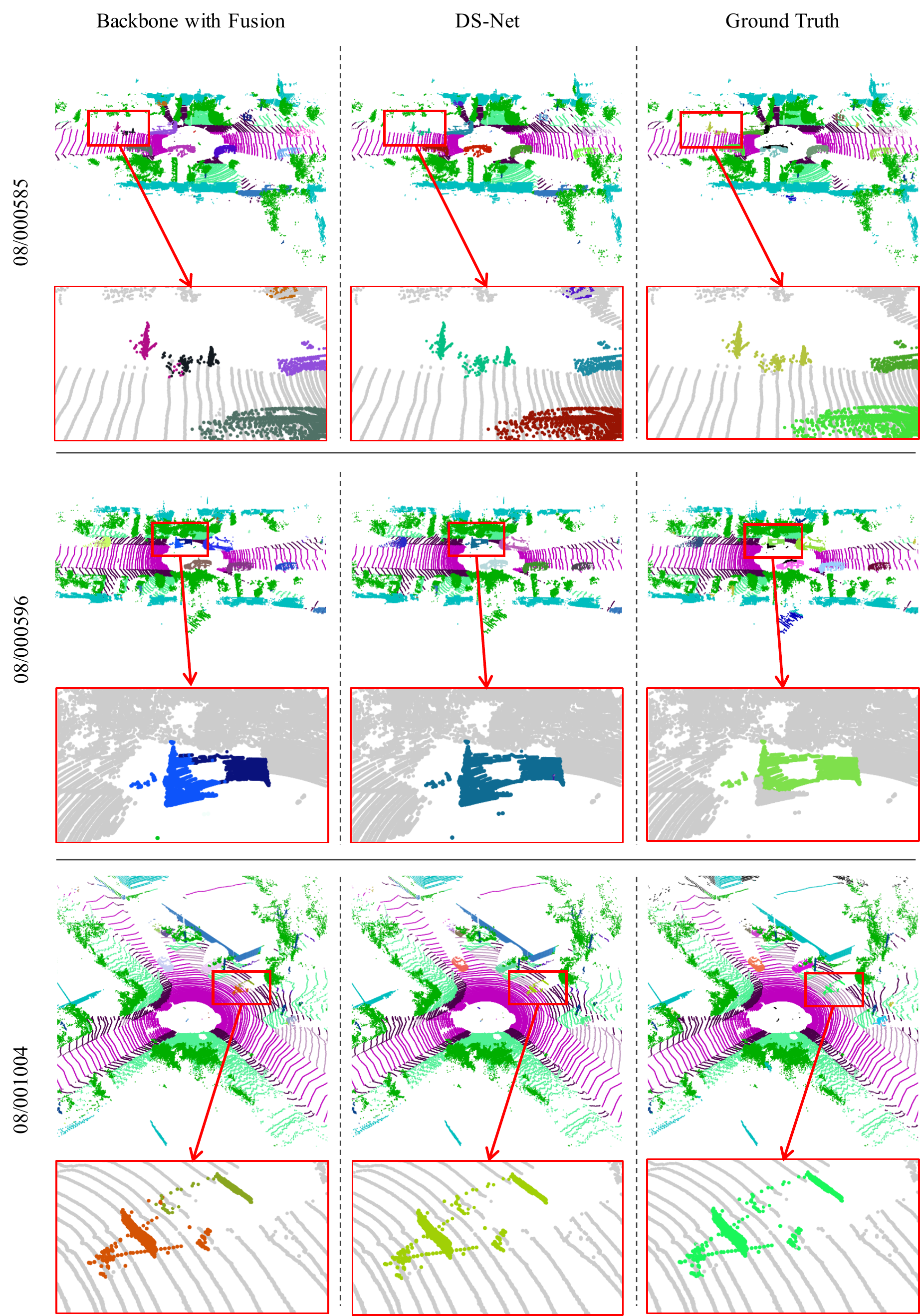}
    \end{center}
    \vspace{-0.6cm}
    \caption{\textbf{Qualitative Comparison of the Backbone and \nickname{} (2).}}
    \label{fig:vis_02}
\end{figure*}

\begin{figure*}[t]
    \vspace{-0.6cm}
    \begin{center}
        \includegraphics[width=0.9\linewidth]{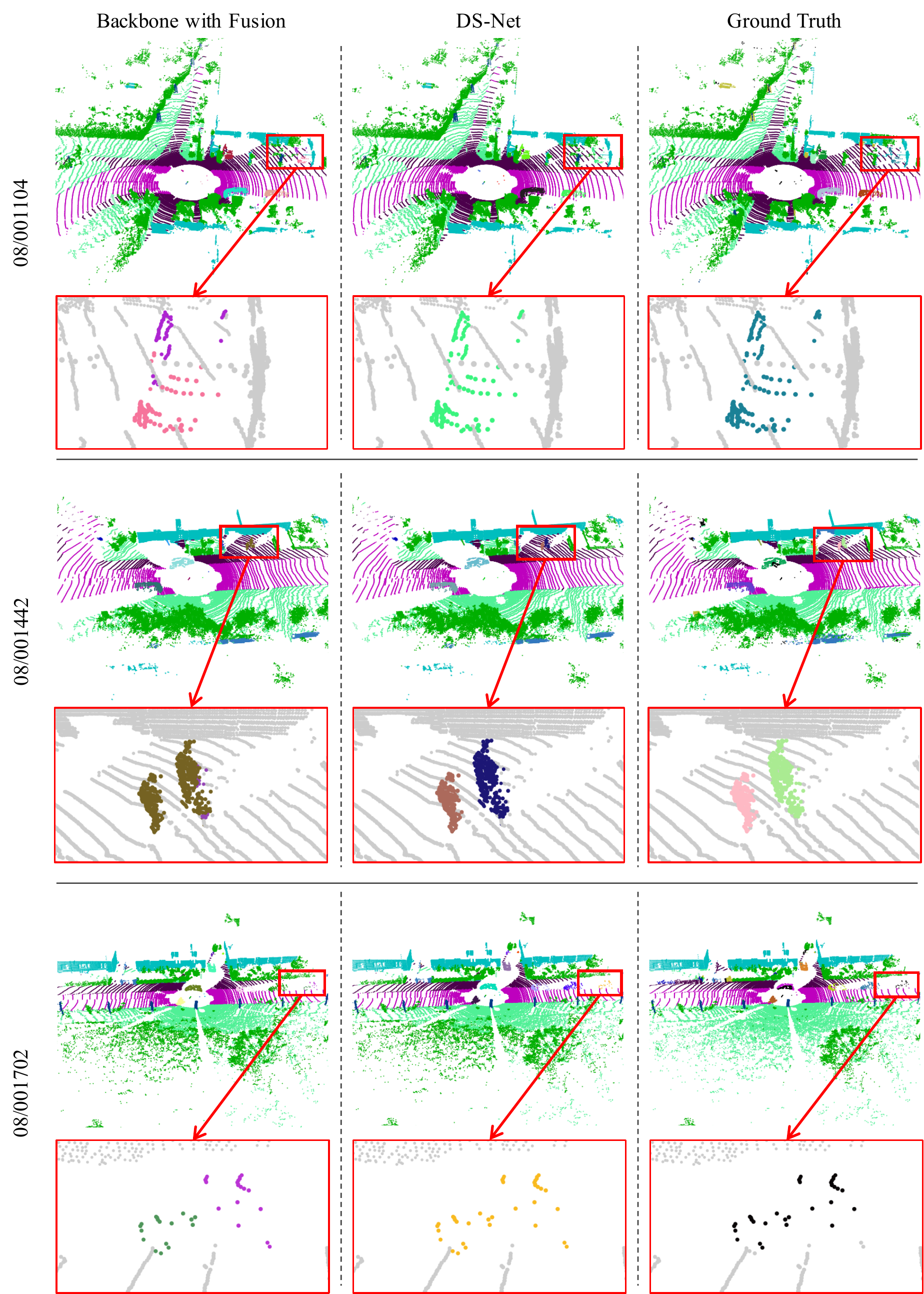}
    \end{center}
    \vspace{-0.6cm}
    \caption{\textbf{Qualitative Comparison of the Backbone and \nickname{} (3).}}
    \label{fig:vis_03}
\end{figure*}

We further show the qualitative comparison of our backbone and \nickname{}.
As shown in Fig. \ref{fig:vis_01}, \ref{fig:vis_02} and \ref{fig:vis_03}, a total of 9 LiDAR frames from the validation set of SemanticKITTI are taken out for visualization.
The left columns are the visualization of the results of our bare backbone with the consensus-driven fusion module.
The middle columns show the results of the \nickname{} and the right columns are the ground truth.
For each frame, a region of interest (as framed in red) is zoomed in to show that our \nickname{} is capable of correctly handling instances with different sizes and densities
while our backbone method tends to either oversegment or undersegment in these complex cases.
Please note that all the \stuff{} classes points are colored following the official definition while the colors of \things{} instances are randomly picked.
To highlight the segmentation results of \things{} classes in the regions of interest, we only color the \things{} class points while the \stuff{} class points are colored as gray.
Since the order of the predicted instance IDs is different from that of the ground truth, the colors of the \things{} instances cannot correspond in predictions and the ground truth.

\end{document}